\documentclass[11pt]{article}

\usepackage[]{acl}

\usepackage{times}
\usepackage{latexsym}
\usepackage{subcaption}
\usepackage{graphicx}
\usepackage{makecell}
\usepackage{paralist}
\usepackage{multirow}
\usepackage[table]{xcolor}

\usepackage[T1]{fontenc}
\usepackage[utf8]{inputenc}

\usepackage{microtype}

\usepackage{inconsolata}

\usepackage{graphicx}

\title{Claim2Vec: Embedding Fact-Check Claims for Multilingual Similarity and Clustering}

\author{Rrubaa Panchendrarajan\\
  Queen Mary University of London \\
  \texttt{r.panchendrarajan@qmul.ac.uk} \\
  \\\And
  Arkaitz Zubiaga \\
  Queen Mary University of London \\
  \texttt{a.zubiaga@qmul.ac.uk}
  }

\begin{document}
\maketitle

\begin{abstract}
Recurrent claims present a major challenge for automated fact-checking systems designed to combat misinformation, especially in multilingual settings. While tasks such as claim matching and fact-checked claim retrieval aim to address this problem by linking claim pairs, the broader challenge of effectively representing groups of similar claims that can be resolved with the same fact-check via claim clustering remains relatively underexplored. To address this gap, we introduce \textit{Claim2Vec}, the first multilingual embedding model optimized to represent fact-check claims as vectors in an improved semantic embedding space. We fine-tune a multilingual encoder using contrastive learning with similar multilingual claim pairs. Experiments on the claim clustering task using three datasets, 14 multilingual embedding models, and 7 clustering algorithms demonstrate that \textit{Claim2Vec} significantly improves clustering performance. Specifically, it enhances both cluster label alignment and the geometric structure of the embedding space across different cluster configurations. Our multilingual analysis shows that clusters containing multiple languages benefit from fine-tuning, demonstrating cross-lingual knowledge transfer.  
\end{abstract}

\section{Introduction}

Automated fact-checking remains challenging due to the rapidly increasing volume of misinformation. A key factor contributing to this growth is the frequent recirculation of previously seen claims \cite{quelle2023lost}. This issue is commonly addressed through tasks such as claim matching—classifying whether a pair of claims is semantically similar—and fact-checked claim retrieval—retrieving similar claims from a fact-check database. However, the recurrence of claims makes both tasks inefficient at scale, motivating more effective approaches such as clustering groups of claims that require the same fact-check. This enables verification or claim retrieval to be performed at the cluster level \cite{multiclaimnet}.  

Claim clustering remains largely underexplored in the literature, with only limited efforts to manually or indirectly verify the existence of such clusters \cite{nielsen2022mumin,hale2024analyzing,shliselberg2024syndy}, primarily due to the lack of suitable datasets. The challenge becomes more pronounced in multilingual settings, which are common in real-world fact-checking scenarios and require language-agnostic methods to identify groups of claims referring to the same factual incident. A recent study introduced \textit{MultiClaimNet} \cite{multiclaimnet}, a collection of three multilingual claim-cluster datasets, and evaluated several multilingual text embedding models  \cite{feng-etal-2022-language,wang2024multilingual,bge-m3,yang2025qwen3} combined with Agglomerative clustering to identify claim clusters. While the authors demonstrated the potential of using multilingual embedding models, these models are primarily designed for general-purpose semantic similarity or retrieval tasks and are not specifically optimized for identifying groups of claims that can be resolved with the same fact-check.   

\begin{figure}[t]
    \centering
\includegraphics[width=0.36\textwidth]{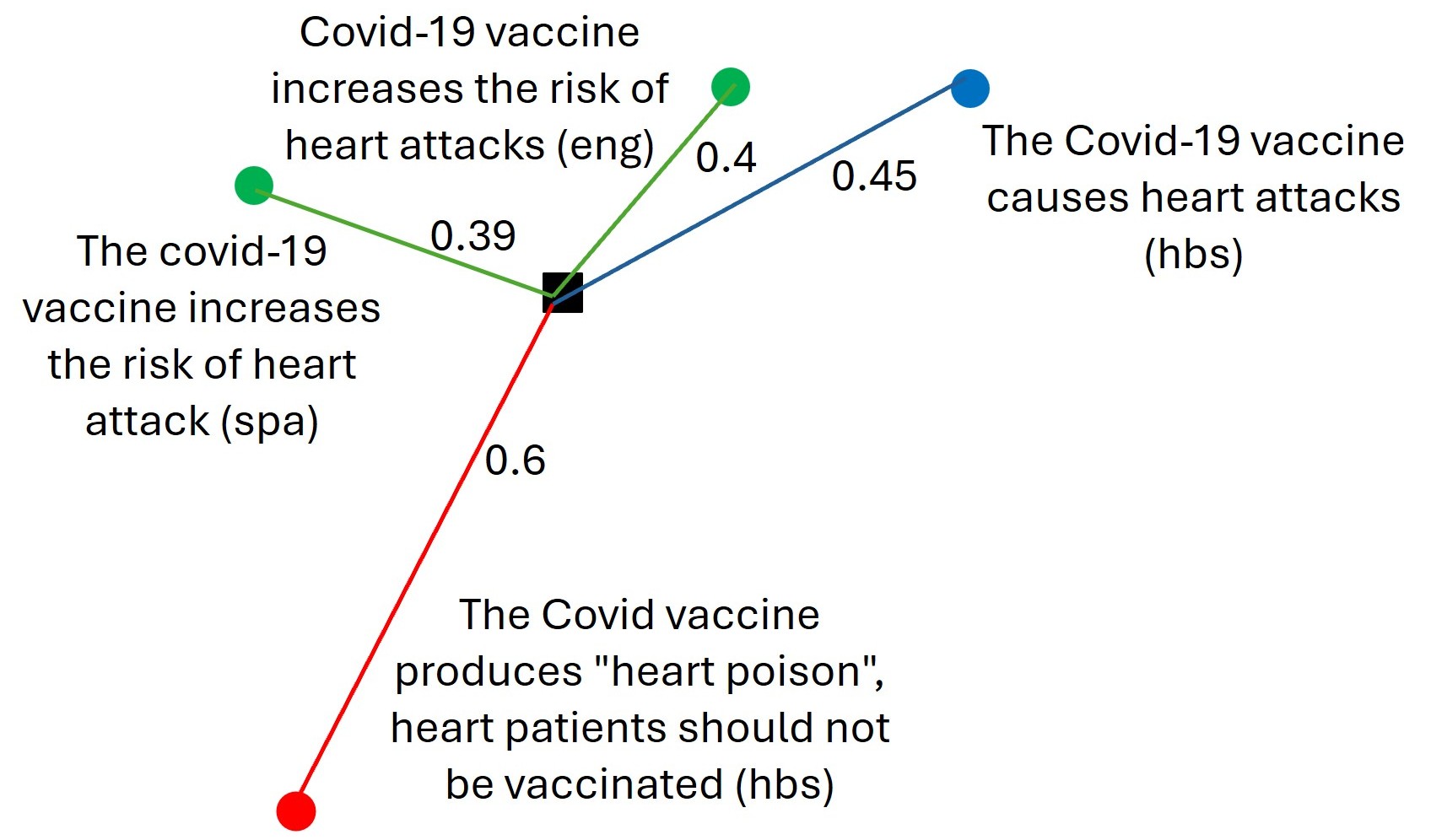}    
    \caption{Sample claims from the same ground-truth cluster in \textit{MultiClaim} Dataset, with \textit{BGE-M3} embeddings projected in 2D, predicted cluster memberships using Agglomerative clustering, and claim language.}
    \label{fig:example_intro}
\end{figure}

Figure \ref{fig:example_intro} illustrates this issue with an example of cluster assignments for claims belonging to the same ground-truth cluster from \textit{MultiClaim} \cite{multiclaimnet}, produced using the multilingual embedding model \textit{BGE-M3} \cite{bge-m3}. The model incorrectly separates the claims into three distinct clusters, primarily due to lexical sensitivity (e.g., ``heart attack” vs. ``heart poison”) and cross-lingual variation (Spanish and English vs. Serbo-Croatian), which increases the embedding distance between semantically equivalent claims. This example highlights the importance of domain-specific embedding representations that more accurately capture multilingual claim similarity.

To address this research gap, we introduce \textit{Claim2Vec}, the first multilingual embedding model specifically designed to represent semantically similar fact-check claims in a shared embedding space. We finetune an existing multilingual embedding model using contrastive learning with similar multilingual claim pairs. The resulting \textit{Claim2Vec} model is evaluated on claim clustering using three multilingual claim-cluster datasets. Experimental results demonstrate that \textit{Claim2Vec} significantly improves clustering performance compared to 14 multilingual embedding models across all three datasets. In particular, \textit{Claim2Vec} enhances both labeling alignment with ground-truth clusters and the geometric structure of the embedding space regardless of the cluster configuration, resulting in improved intra-cluster cohesion and greater inter-cluster separation. We further show that \textit{Claim2Vec} substantially corrects the incorrect cluster splits produced by its pretrained counterpart, \textit{BGE-M3} \cite{bge-m3}, by learning more robust cross-lingual semantic representations. Moreover, our multilingual analysis reveals that ground-truth clusters containing claims written in multiple languages benefit most from the fine-tuning process, indicating effective cross-lingual knowledge transfer.

We make the following contributions.
\begin{compactitem}
\item We introduce \textit{Claim2Vec}\footnote{\url{https://huggingface.co/Rrubaa/claim2vec}}, the first multilingual embedding model tailored for fact-check claims, enabling improved representation of semantically similar claims across languages.
\item We apply contrastive learning with multilingual claim pairs, demonstrating enhanced semantic alignment and cross-lingual knowledge transfer.
\item We conduct extensive experiments across three datasets, seven clustering algorithms, and 14 multilingual embedding models to comprehensively evaluate the effectiveness and robustness of \textit{Claim2Vec}.
\end{compactitem}

\section{Related Work}

\paragraph{\textbf{Multilingual Text Embedding.}} While language-specific text representations have been widely explored, extending them to a universal multilingual embedding space remains a challenging problem. Early work such as LASER \cite{artetxe2019massively} addressed this by training an LSTM-based encoder–decoder architecture for machine translation on large parallel corpora. Subsequent research adopted transformer-based architectures, leading to models such as LaBSE \cite{feng-etal-2022-language}, which further improved multilingual representation learning. A major shift in this direction came with the introduction of contrastive pretraining, where models are first trained with weak supervision on large-scale parallel or multilingual corpora and later fine-tuned on smaller, high-quality labeled datasets. Models such as Multilingual E5 \cite{wang2024multilingual} demonstrated the effectiveness of contrastive learning for multilingual embeddings. This multi-stage training paradigm was subsequently extended to retrieval-focused models such as BGE-M3 \cite{bge-m3}. More recent work has explored leveraging large language models to eliminate the initial pretraining stage \cite{wang2024improving,yang2025qwen3}. These developments have resulted in a wide range of multilingual text embedding models. In this work, we compare the best-performing multilingual text embedding (MTE) models evaluated on the MTEB benchmark, which covers 250 languages and 500 tasks \cite{muennighoff2023mteb,enevoldsen2025mmteb}.

\paragraph{\textbf{Claim Matching.}} As closely related task, claim matching typically involves determining whether a pair of claims expresses the same or similar information and is commonly formulated as a binary classification problem. A widely adopted approach is to fine-tune multilingual transformer-based models on labeled datasets for claim similarity classification \cite{bouziane2020team,mansour2022did,mansour2023not}. However, the extent to which such models can produce effective claim representations for downstream tasks such as clustering and retrieval remains largely unexplored.  

\paragraph{\textbf{Claim Clustering.}} Only few studies tackled claim clustering, which focused mainly on either manually or indirectly verifying the existence of claim clusters using unsupervised techniques such as Agglomerative clustering \cite{kazemi2021claim,hale2024analyzing}, HDBSCAN \cite{nielsen2022mumin}, and topic-based grouping approaches \cite{smeros2021sciclops,shliselberg2024syndy}. More recently, \citet{multiclaimnet} introduced a collection of three claim-cluster datasets constructed using a combination of manual and LLM-based annotation strategies. Using these datasets, the authors evaluated multiple clustering algorithms and multilingual embedding models. Their results show that Agglomerative clustering applied to embeddings from BGE-M3 \cite{bge-m3}, a retrieval-oriented embedding model, performs best in the task, which we adopt as one of our baselines.

\begin{figure*}[t]
\centering    
\includegraphics[width=0.79\textwidth]{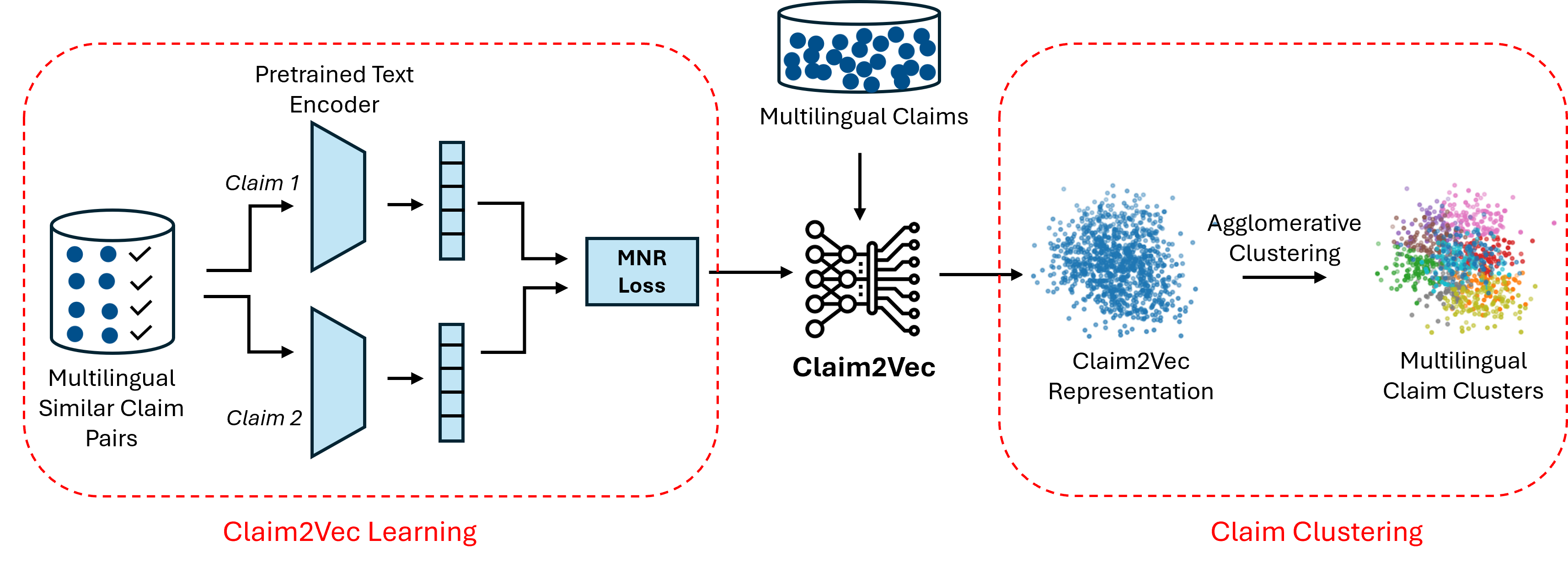}
\caption{Methodology of \textit{Claim2Vec} learning and claim clustering}
\label{fig:methodology}
\end{figure*}

\section{Methodology}

Figure \ref{fig:methodology} illustrates the overall methodology, which consists of two main stages: \textit{Claim2Vec} representation learning and claim clustering.
% The following subsections describe each stage in detail.

\subsection{Claim2Vec Learning}

\subsubsection{Training Data}\label{sec:training_data}
To project multilingual claims into a vector space where semantically similar claims are positioned close to one another, we rely on training data consisting of multilingual claim pairs annotated for semantic similarity. For this purpose, we utilize the claim pairs originally employed to construct the ground-truth clusters in the \textit{MultiClaimNet} datasets \cite{multiclaimnet}. This collection comprises three ground-truth claim-cluster datasets  --\textit{ClaimCheck, ClaimMatch} and \textit{MultiClaim}-- derived from manually or automatically annotated similar claim pairs (see Section~\ref{sec:datasets}).

The largest of these datasets, \textit{MultiClaim}, was constructed by first retrieving nearest neighbors from a raw collection of fact-checked claims to form candidate claim pairs. A total of 185K candidate pairs were then annotated using three large language models. Pairs labeled as similar were retained, and clusters were automatically formed based on transitive similarity links among claims (e.g., if pairs A–B and B–C are similar, then A, B, and C are grouped into the same cluster). We use \textit{MultiClaim} for both \textit{Claim2Vec} training and clustering evaluation. The dataset comprises 85.3K fact-checked claims written in 78 languages and grouped into 30.9K clusters. The dataset construction assumes that each claim expresses a single factual statement. To ensure consistency with the training and evaluation data, we adopt the same assumption in our work.  

For a robust evaluation setting, we partition the claims in \textit{MultiClaim} into two disjoint topic groups such that no overlapping topics appear across the training and test sets, while preserving cluster integrity in the test partition. To achieve this, all claims belonging to the same ground-truth cluster are first merged into a single document. We then apply Latent Dirichlet Allocation (LDA) to divide the dataset into two coherent topic groups. Topic group 1 primarily covers global political issues and is used for training. From this group, we leverage the 28K similar claim pairs used to construct the clusters belonging to this group to train \textit{Claim2Vec}. Topic group 2 predominantly focuses on COVID-19–related discussions, and we use the ground-truth clusters belonging to this group for testing claim clustering. These partitions are denoted as \textit{MultiClaim-Train} and \textit{MultiClaim-Test}, respectively. The \textit{MultiClaim-Train} partition consists of 43K multilingual claims, representing nearly half of the 85.3K claims in \textit{MultiClaim}, and covers 71 languages. \textit{MultiClaim} Train and Test partitions are available online\footnote{\url{https://zenodo.org/records/19494468}}. Additional details of the two topic groups are provided in Appendix~\ref{app:multilclaim_topics}.

\subsubsection{Contrastive Learning}
Using the training partition, we fine-tune a pretrained multilingual embedding model to learn claim representations such that semantically similar claims are positioned close to one another in the embedding space, while dissimilar claims are pushed farther apart to facilitate clustering. We do not incorporate the dissimilar pairs annotated during the construction of \textit{MultiClaim}, as these pairs were derived from nearest-neighbor retrieval and therefore constitute only hard negatives. Instead, we use the 28K similar claim pairs from topic group 1 as positive training instances. %As the base encoder, we adopt \textit{BGE-M3} \cite{bge-m3}, a retrieval-based multilingual embedding model that has demonstrated strong performance on claim clustering tasks \cite{multiclaimnet}. 

We finetuned \textit{BGE-M3} \cite{bge-m3}, a retrieval-based multilingual embedding model using the Multiple Negatives Ranking Loss (MNRL) \cite{henderson2017efficient}, a contrastive learning objective that leverages in-batch training. For each positive pair within a batch, all other claims in the same batch are treated as implicit negative samples. A softmax function is applied over the similarity scores to maximize the similarity of the true pairs while minimizing similarity with in-batch negatives, thereby structuring the embedding space for effective clustering. Although in-batch negatives may introduce occasional false negatives, their impact is limited in our setting. The \textit{MultiClaim} dataset has an average cluster size of 2.6, with most clusters containing only two claims. Consequently, the probability that two semantically equivalent claims appear in the same batch and are mistakenly treated as negatives is low. We employ cosine similarity as the metric within the loss function and fine-tune the model for one epoch using a batch size of 32 and a learning rate of $1x10^{-5}$. We refer to the finetuned model as \textit{Claim2Vec}.

\begin{figure}[t]
    \centering
    \begin{subfigure}[b]{0.23\textwidth}
    \includegraphics[width=\textwidth]{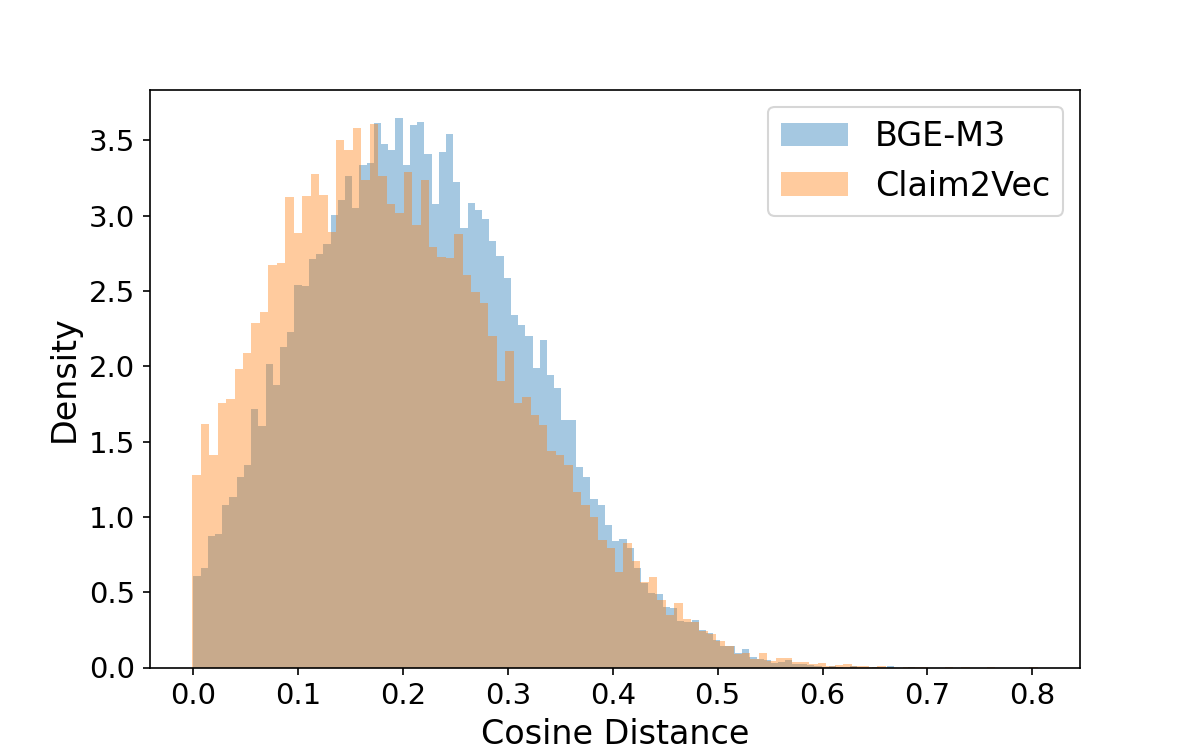}
    \caption{Positive Pairs}
    \label{fig:positive_pair_distance_comparison}   
    \end{subfigure}  
    \begin{subfigure}[b]{0.23\textwidth}
    \includegraphics[width=\textwidth]{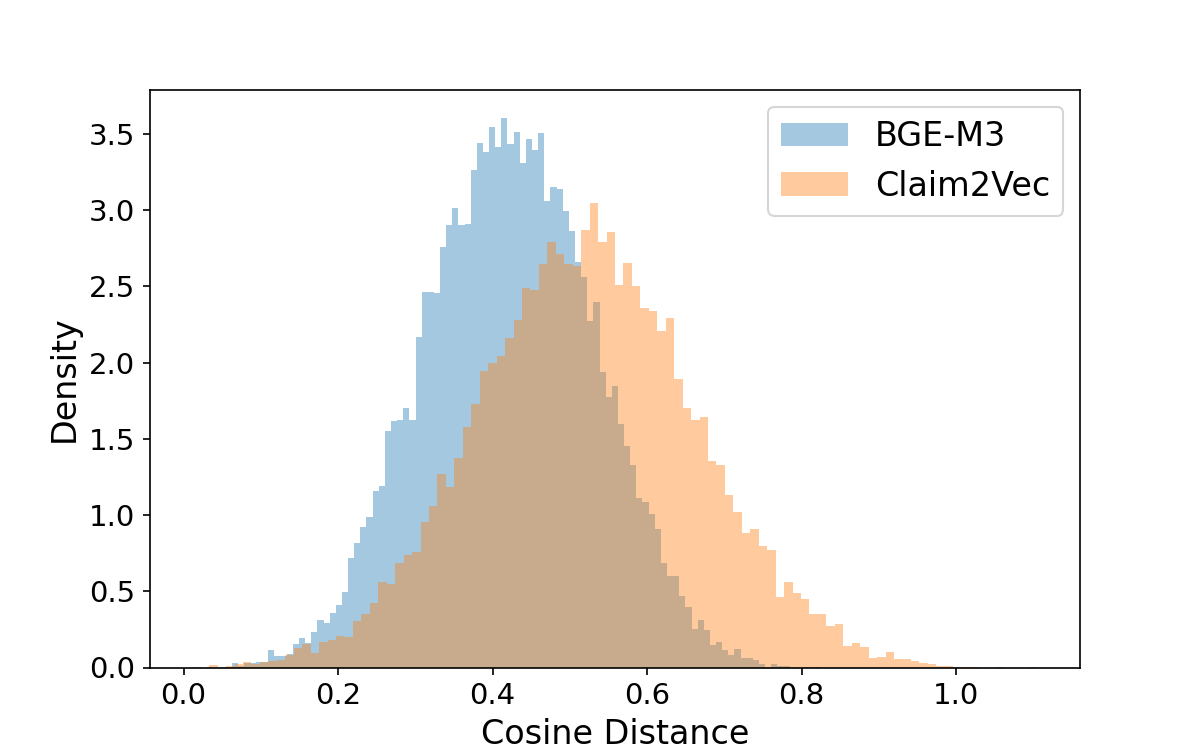}
    \caption{Negative Pairs}
    \label{fig:negative_pair_distance_comparison}   
    \end{subfigure}
    \caption{Positive and negative pairs' cosine distance distribution - \textit{BGE-M3} vs \textit{Claim2Vec}}
    \label{fig:pair_distance_comparison}
\end{figure}

Figure \ref{fig:positive_pair_distance_comparison} shows cosine distance distribution for similar claim pairs, while Figure \ref{fig:negative_pair_distance_comparison} shows the distribution for dissimilar claim pairs, obtained using \textit{BGE-M3} and \textit{Claim2Vec}. Although dissimilar claim pairs from topic group 1 were not used during training, we analyzed their distance distributions to assess the effect of fine-tuning. Negative pairs were derived by filtering the 185K annotated claim pairs from \textit{MultiClaim} and assigning them to topic group 1 or 2 using the same topic model. \textit{Claim2Vec} brings similar claim pairs closer together in the embedding space, shifting their cosine distance distribution to the left, with the mean decreasing from 0.2185 to 0.1993. In contrast, the distribution of dissimilar pairs shifts to the right, with the mean increasing from 0.4219 to 0.522, indicating greater separation. Notably, these negative pairs were not observed during training. This consistent divergence demonstrates that fine-tuning effectively restructures the embedding space to better capture semantic similarity between claims.

\subsection{Claim Clustering}\label{sec:clustering}
We first encode claims into dense vector representations using \textit{Claim2Vec}, and then apply a clustering algorithm to group claims that can be verified together. We adopt Agglomerative Clustering \cite{mullner2011modern}, which has been shown to perform effectively in \textit{MultiClaimNet} \cite{multiclaimnet}, outperforming other clustering algorithms evaluated. Agglomerative clustering is a bottom-up hierarchical method that initially treats each data point as an individual cluster and iteratively merges the closest cluster pairs until all points are combined into a single cluster. Here, the distance threshold provided as a hyperparameter determines the cut point in the resulting dendrogram and thus defines the final clustering configuration.

In \cite{multiclaimnet}, the distance threshold was manually selected to achieve strong performance across all three datasets. However, each dataset may exhibit different distance distributions, making a single manual threshold suboptimal, and manual selection is impractical without ground-truth clusters. We instead automate threshold selection by maximizing the Silhouette Score \cite{rousseeuw1987silhouettes}, which measures intra-cluster cohesion and inter-cluster separation. We vary the distance threshold from 0.5 to 1.5 (step 0.05) and select the threshold yielding the highest Silhouette Score. To avoid local optimum, we further refine the search by evaluating 10 additional candidate cuts on both sides of the best candidate in the dendrogram. For \textit{MultiClaim-Test}, the threshold is estimated as the average over five random subsets of 10,000 records due to computational limits. Empirically, this automated strategy consistently improves clustering performance across all three datasets, irrespective of the embedding model or clustering algorithm used, compared to the baseline results in \citet{multiclaimnet}.

\begin{table}[t]
\scriptsize
\centering
\begin{tabular}{llllll}\hline
Dataset & 
Clusters & 
Claims & 
Avg & 
Max & 
Languages \\ 
\hline
ClaimCheck  & 197         & 1187      & 6.03                                                              & 28                                                             & 22                                                                              \\
ClaimMatch & 192         & 1171      & 6.1                                                               & 35                                                             & 36                                                                            \\
MultiClaim-Test & 15,988       & 42.4K     & 2.65                                                              & 31                                                             & 64                              \\ \hline                                           
\end{tabular}
\caption{Statistics of cluster datasets}\label{tab:data_stats}
\end{table}

\section{Experiment Setup}
%We analyze the performance of the \textit{Claim2Vec} model on the claim clustering task by comparing it with various multilingual embedding models and clustering algorithms across different datasets. 

\subsection{Datasets}\label{sec:datasets}
We evaluate the \textit{Claim2Vec} model on claim clustering using \textit{MultiClaimNet} datasets: \textit{ClaimCheck}, \textit{ClaimMatch}, and \textit{MultiClaim} (see Table \ref{tab:data_stats} for statistics). \textit{ClaimCheck} and \textit{ClaimMatch} are relatively small and derived from existing claim-matching resources annotated for semantic similarity. Although they are comparable in size (number of claims) and density (average cluster size), \textit{ClaimCheck} is comparatively easier, as the original claim-matching dataset was partially generated through automated similarity-based annotation that produce clearer cluster separation. In contrast, \textit{ClaimMatch} presents greater ambiguity between clusters. The largest dataset, \textit{MultiClaim}, was created using the automated methodology described in Section~\ref{sec:training_data}. For clustering evaluation, we use the \textit{MultiClaim-Test} partition. It contains a significantly larger number and linguistically more diverse set of claims, increasing semantic variability and cross-lingual complexity, making clustering substantially harder despite small average cluster sizes.

\begin{table}[t]
\centering
\scriptsize
\renewcommand{\arraystretch}{0.9}
\begin{tabular}{p{4cm}ll} \hline
Model                                          &  Parameters  & Embedding \\ \hline
Distiluse-base-multilingual-cased           & 135M & 512       \\
Paraphrase-multilingual-MiniLM-L12          & 118M & 768       \\
Paraphrase-multilingual-mpnet-base          & 278M & 768       \\
Gte-multilingual-base                          & 305M & 768      \\
All-roberta-large-v1                           & 355M & 1024      \\
LaBSE                                          & 471M & 768       \\
KaLM-mini-instruct-v1.5 & 494M & 896       \\
Multilingual-e5-large-instruct                 & 560M & 1024      \\
Bge-m3                                         & 567M & 1024      \\ \hline
Gte-Qwen2-1.5B-instruct                        & 1B   & 1536      \\
MiniCPM-Embedding                              & 2.4B & 2304      \\
E5-mistral-7b-instruct                         & 7B   & 4096      \\
Gte-Qwen2-7B-instruct                          & 7B   & 3584      \\
Bge-multilingual-gemma2                        & 9B   & 3584    \\ \hline 
\end{tabular}
\caption{Multilingual text embedding models}\label{tab:sentence_embedding}
\end{table}

\begin{table*}[t]
\centering
\scriptsize
\renewcommand{\arraystretch}{0.9}
\begin{tabular}{lccccccccc}
\hline
& \multicolumn{3}{c}{ClaimCheck} 
& \multicolumn{3}{c}{ClaimMatch} 
& \multicolumn{3}{c}{MultiClaim-Test} \\
\hline
Approach 
& \#Clusters & ARI & AMI 
& \#Clusters & ARI & AMI 
& \#Clusters & ARI & AMI \\
\hline
HDBSCAN & 175 & 0.847 & 0.912 & 172 & 0.685 & 0.824 & 7765  & 0.007 & 0.416 \\
Agglomerative & \textbf{200} & \textbf{0.906} & \textbf{0.955} 
              & \textbf{210} & \textbf{0.762} & \textbf{0.886} 
              & \textbf{16344} & \textbf{0.625} & \textbf{0.757} \\
Birch & 332 & 0.789 & 0.867 & 333 & 0.684 & 0.822 & 24804 & 0.449 & 0.567 \\
Optics & 303 & 0.093 & 0.552 & 289 & 0.064 & 0.486 & 10323 & 0.001 & 0.227 \\
MeanShift & 354 & 0.836 & 0.879 & 329 & 0.589 & 0.815 & -- & -- & -- \\
Affinity Propagation & 177 & 0.882 & 0.936 & 151 & 0.740 & 0.850 & -- & -- & -- \\
Leiden Community Detection & 155 & 0.815 & 0.899 & 146 & 0.692 & 0.837 & 11859 & 0.371 & 0.627 \\
\hline
\end{tabular}
\caption{Clustering performance of different clustering algorithms.}
\label{tab:clustering_results}
\end{table*}

\begin{table*}[t]
\centering
\scriptsize
\renewcommand{\arraystretch}{0.9}
\begin{tabular}{lccccccccc}
\hline
 & \multicolumn{3}{c}{ClaimCheck} 
 & \multicolumn{3}{c}{ClaimMatch} 
 & \multicolumn{3}{c}{MultiClaim-Test} \\ \hline
Embedding Model 
 & ARI & AMI & SS 
 & ARI & AMI & SS 
 & ARI & AMI & SS \\
\hline
Distiluse-base-multilingual-cased-v1 
 & 0.765 & 0.874 & 0.237 
 & 0.509 & 0.693 & 0.187 
 & 0.413 & 0.566 & 0.191 \\

Paraphrase-multilingual-MiniLM-L12-v2 
 & 0.814 & 0.896 & 0.290 
 & 0.660 & 0.797 & 0.229 
 & 0.475 & 0.622 & 0.222 \\

Paraphrase-multilingual-mpnet-base-v2 
 & 0.837 & 0.916 & 0.332 
 & 0.701 & 0.833 & 0.261 
 & 0.540 & 0.673 & 0.246 \\

Gte-multilingual-base 
 & 0.882 & 0.929 & 0.285 
 & 0.613 & 0.797 & 0.211 
 & 0.569 & 0.698 & 0.237 \\

All-roberta-large-v1 
 & 0.409 & 0.633 & 0.227 
 & 0.337 & 0.528 & 0.220 
 & 0.262 & 0.433 & 0.189 \\

LaBSE 
 & 0.751 & 0.871 & 0.231 
 & 0.456 & 0.666 & 0.185 
 & 0.511 & 0.623 & 0.198 \\

KaLM-mini-instruct-v1.5 
 & 0.596 & 0.786 & 0.219 
 & 0.391 & 0.615 & 0.193 
 & 0.479 & 0.591 & 0.206 \\

Multilingual-e5-large-instruct 
 & 0.720 & 0.871 & 0.224 
 & 0.440 & 0.677 & 0.195 
 & 0.518 & 0.631 & 0.216 \\

BGE-M3 
 & 0.845 & 0.929 & 0.264 
 & 0.644 & 0.820 & 0.204 
 & 0.610 & 0.726 & 0.229 \\

Gte-Qwen2-1.5B-instruct 
 & \textbf{0.907} & 0.946 & 0.274 
 & 0.553 & 0.762 & 0.204 
 & 0.532 & 0.656 & 0.222 \\

MiniCPM-Embedding 
 & 0.636 & 0.814 & 0.228 
 & 0.406 & 0.627 & 0.194 
 & 0.448 & 0.576 & 0.204 \\

E5-mistral-7b-instruct 
 & 0.819 & 0.904 & 0.233 
 & 0.455 & 0.692 & 0.202 
 & 0.508 & 0.623 & 0.211 \\

Gte-Qwen2-7B-instruct 
 & 0.856 & 0.919 & 0.243 
 & 0.506 & 0.730 & 0.183 
 & 0.540 & 0.664 & 0.209 \\

Bge-multilingual-gemma2 
 & 0.608 & 0.799 & 0.223 
 & 0.418 & 0.660 & 0.202 
 & 0.479 & 0.604 & 0.206 \\ \hline

Claim2Vec 
 & 0.906 & \textbf{0.955} & \textbf{0.354} 
 & \textbf{0.762} & \textbf{0.886} & \textbf{0.287} 
 & \textbf{0.626} & \textbf{0.758} & \textbf{0.284} \\
\hline
\end{tabular}
\caption{Clustering Performance of Multilingual Embedding Models.}
\label{tab:clustering_results_embedding_models}
\end{table*}

\begin{table*}[!t]
\centering
\scriptsize
\renewcommand{\arraystretch}{0.9}
\begin{tabular}{llccccccc}
\hline
Dataset & Embedding Model & \#Clusters & ARI & AMI & HOM & CMP & VM & SS \\
\hline
\multirow{2}{*}{ClaimCheck (197)}
& BGE-M3 
& \textbf{209} 
& 0.845 
& 0.929 
& 0.985 
& 0.961 
& 0.973 
& 0.264 \\

& Claim2Vec 
& 200 
& \textbf{0.906} 
& \textbf{0.955} 
& \textbf{0.988} 
& \textbf{0.977} 
& \textbf{0.982} 
& \textbf{0.354} \\
\hline

\multirow{2}{*}{ClaimMatch (192)}
& BGE-M3 
& \textbf{258} 
& 0.644 
& 0.820 
& \textbf{0.974} 
& 0.902 
& 0.936 
& 0.204 \\

& Claim2Vec 
& 210 
& \textbf{0.762} 
& \textbf{0.886} 
& 0.972 
& \textbf{0.941} 
& \textbf{0.956} 
& \textbf{0.287} \\
\hline

\multirow{2}{*}{MultiClaim-Test (15988)}
& BGE-M3 
& \textbf{17922} 
& 0.610 
& 0.726 
& \textbf{0.976} 
& 0.967 
& 0.971 
& 0.229 \\

& Claim2Vec 
& 16338 
& \textbf{0.626} 
& \textbf{0.758} 
& 0.972 
& \textbf{0.975} 
& \textbf{0.973} 
& \textbf{0.284} \\
\hline
\end{tabular}
\caption{Clustering Comparison between \textit{BGE-M3} and \textit{Claim2Vec}. }
\label{tab:bge_vs_claim2vec}
\end{table*}

\subsection{Clustering Algorithms}
We compare seven clustering algorithms, which do not require the number of clusters to be specified as the hyperparameters: Agglomerative clustering \cite{mullner2011modern}, HDBSCAN \cite{mcinnes2017hdbscan}, Affinity Propagation \cite{dueck2009affinity}, Birch \cite{zhang1996birch}, MeanShift \cite{comaniciu2002mean}, Optics \cite{ankerst1999optics}, Leiden Community Detection \cite{traag2019louvain}. Hyperparameters are automatically selected by maximizing the Silhouette Score (refer to Appendix \ref{app:hyperparamters}). For density-based methods (HDBSCAN and BIRCH), embeddings are first reduced with UMAP \cite{mcinnes2018umap}, with the optimal dimensionality also chosen via Silhouette Score. %Appendix \ref{app:hyperparamters} details the hyperparameter search space and the optimal configurations identified for each dataset.

\subsection{Baseline Models}
Table \ref{tab:sentence_embedding} lists the 14 multilingual text embedding models used for comparison with \textit{Claim2Vec}. These models vary substantially in scale, ranging from 135M to 9B parameters, and were selected based on their performance on the multilingual text embedding benchmark (MTEB).\footnote{https://huggingface.co/spaces/mteb/leaderboard}

\subsection{Evaluation Metrics}
For clustering evaluation \cite{pauletic2019overview}, we report the following metrics along with the Silhouette Score: Adjusted Rand Index (ARI), which quantifies the similarity between predicted and ground-truth clusters; Adjusted Mutual Information (AMI), which measures the mutual information shared between predicted and ground-truth clusters; Homogeneity (HMG), the fraction of predicted cluster members that belong to a single ground-truth cluster; Completeness (CMP), the fraction of instances of a ground-truth cluster assigned to the same predicted cluster; and V-Measure (VM), the harmonic mean of homogeneity and completeness. ARI and Silhouette Score range from $-1$ to $1$ and other metrics range from $0$ to $1$, with higher values indicate better clustering.

\section{Results}

\subsection{Clustering Algorithms}

Table \ref{tab:clustering_results} shows results for the different clustering algorithms\footnote{Due to high computational cost, we do not report results for MeanShift and Affinity Propagation on \textit{MultiClaim-Test}.}. We observe that Agglomerative Clustering demonstrates the most consistent performance. While HDBSCAN performs competitively on the smaller datasets, its performance degrades substantially as the number of data points increases. In contrast, the community detection approach scales better to the larger dataset compared to HDBSCAN; however, it produces a high number of mismerge errors, resulting in significantly fewer predicted clusters than the ground-truth count. This behavior suggests that the method tends to form coarse-grained clusters of claims that share broadly similar topics, making it difficult to recover fine-grained, semantically precise groupings.

\subsection{Multilingual Text Embedding Models}
Table \ref{tab:clustering_results_embedding_models} reports the clustering performance of the 14 multilingual embedding models alongside \textit{Claim2Vec}, using Agglomerative Clustering. Except \textit{GTE-Qwen2-1.5B-Instruct}, which achieves the same ARI as \textit{Claim2Vec} on \textit{ClaimCheck}, \textit{Claim2Vec} consistently outperforms all compared multilingual models across the three datasets in terms of ARI, AMI, and Silhouette Score. The substantial improvements across these metrics indicate that \textit{Claim2Vec} more effectively captures semantic similarity between claims, leading to better cluster assignments (as reflected by ARI and AMI) as well as improved geometric separation in the embedding space (as reflected by the Silhouette Score).

\subsection{Impact of Finetuning}
We assess the impact of fine-tuning by comparing the performance of the base pretrained encoder, \textit{BGE-M3}, against the fine-tuned model, \textit{Claim2Vec}. 

\subsubsection{Clustering Performance}
Table \ref{tab:bge_vs_claim2vec} compares clustering performance of \textit{BGE-M3} and \textit{Claim2Vec}. Except for Homogeneity, \textit{Claim2Vec} consistently achieves higher scores across ARI, AMI, and Silhouette Score on all three datasets, and significantly improves Completeness on \textit{ClaimMatch}. This shows that fine-tuning enhances semantic representation, yielding more accurate cluster assignments and better geometric separation. Notably, \textit{BGE-M3} tends to over-segment, producing more clusters than ground truth, whereas \textit{Claim2Vec} reduces such invalid splits, generating clusters that align more closely with the true labels. This behavior also explains why both models achieve similar Homogeneity scores which is insensitive over-segmentation.
 
\begin{figure*}[t]
    \centering
    \begin{subfigure}[b]{0.3\textwidth}
    \includegraphics[width=\textwidth]{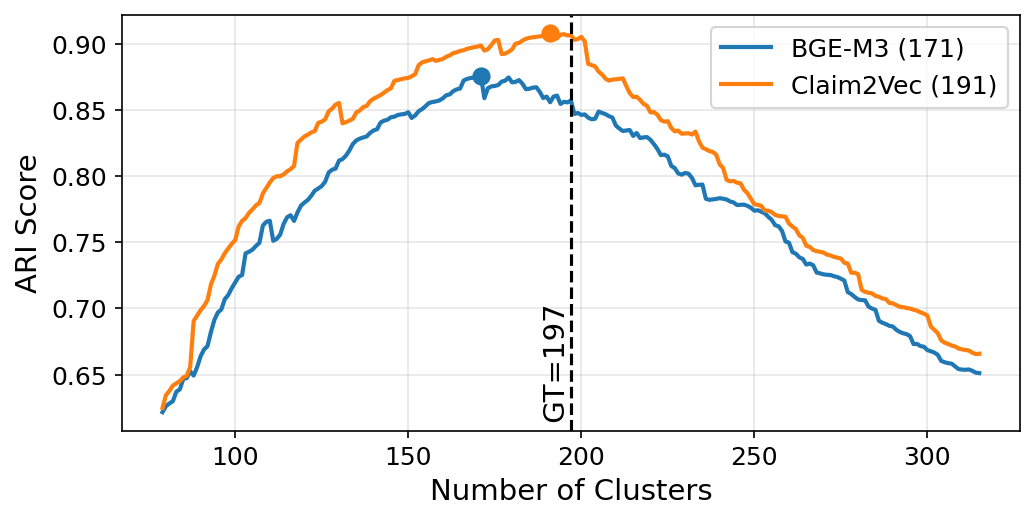}
    \caption{ARI - \textit{ClaimCheck}}
    \label{fig:ari_comparison_FactCheck}   
    \end{subfigure}  
    \begin{subfigure}[b]{0.3\textwidth}
    \includegraphics[width=\textwidth]{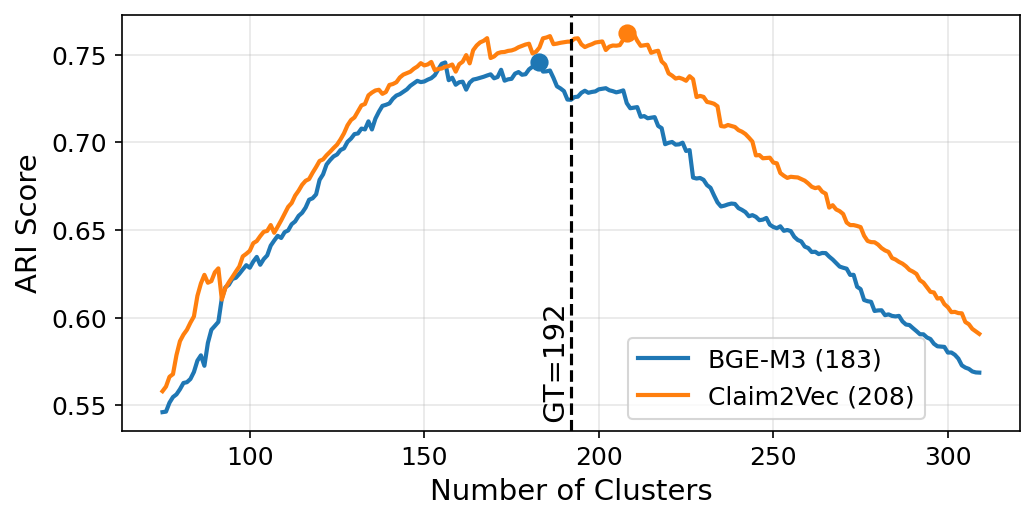}
    \caption{ARI - \textit{ClaimMatch}}
    \label{fig:ari_comparison_ClaimMatching}   
    \end{subfigure}
    \begin{subfigure}[b]{0.3\textwidth}
    \includegraphics[width=\textwidth]{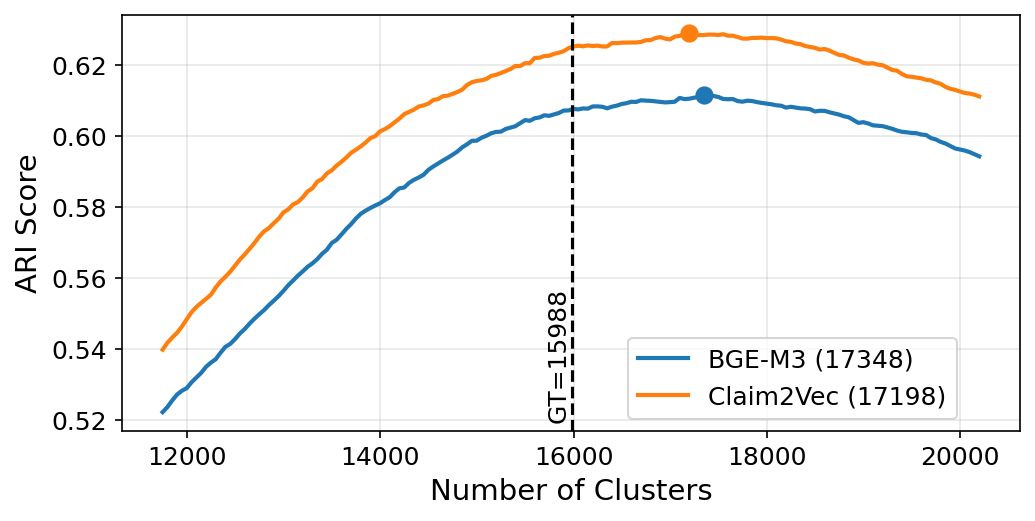}
    \caption{ARI - \textit{MultiClaim-Test}}
    \label{fig:ari_comparison_MultiClaim_1}   
    \end{subfigure}

    \begin{subfigure}[b]{0.3\textwidth}
    \includegraphics[width=\textwidth]{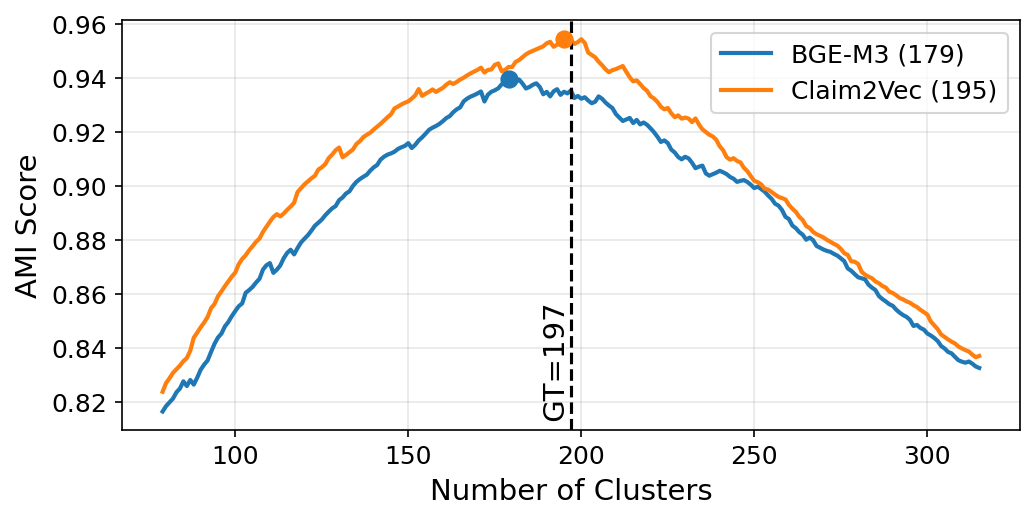}
    \caption{AMI - \textit{ClaimCheck}}
    \label{fig:ami_comparison_FactCheck}   
    \end{subfigure}  
    \begin{subfigure}[b]{0.3\textwidth}
    \includegraphics[width=\textwidth]{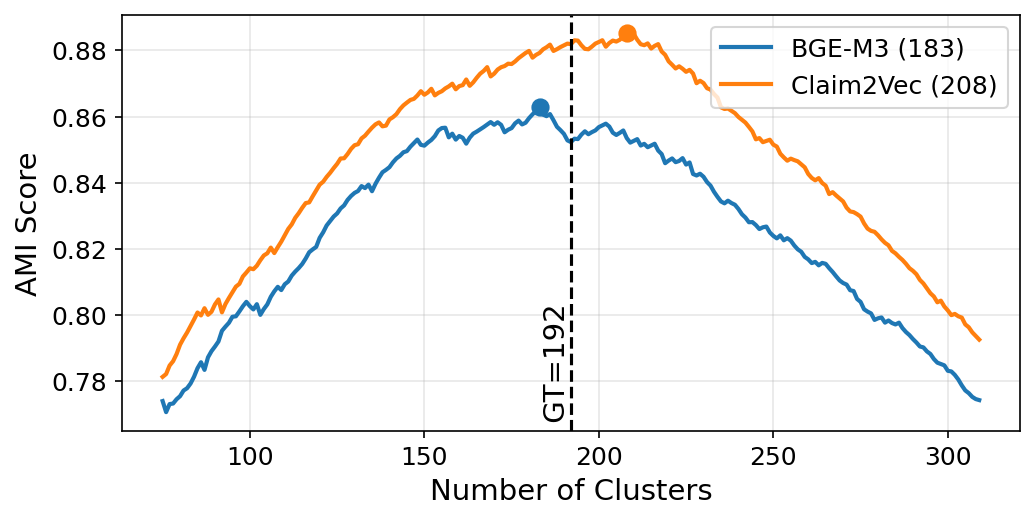}
    \caption{AMI - \textit{ClaimMatch}}
    \label{fig:ami_comparison_ClaimMatching}   
    \end{subfigure}
    \begin{subfigure}[b]{0.3\textwidth}
    \includegraphics[width=\textwidth]{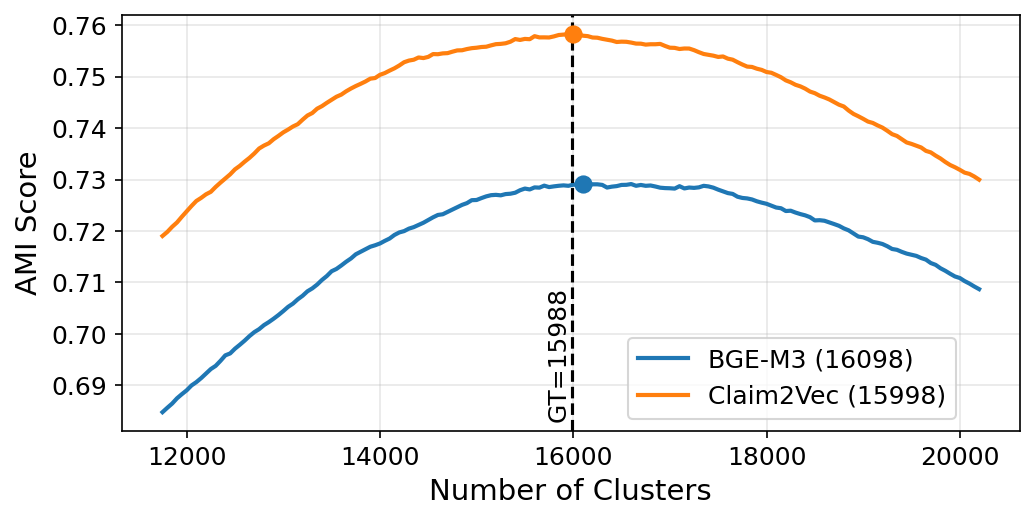}
    \caption{AMI - \textit{MultiClaim-Test}}
    \label{fig:ami_comparison_MultiClaim_1}   
    \end{subfigure}

    \begin{subfigure}[b]{0.3\textwidth}
    \includegraphics[width=\textwidth]{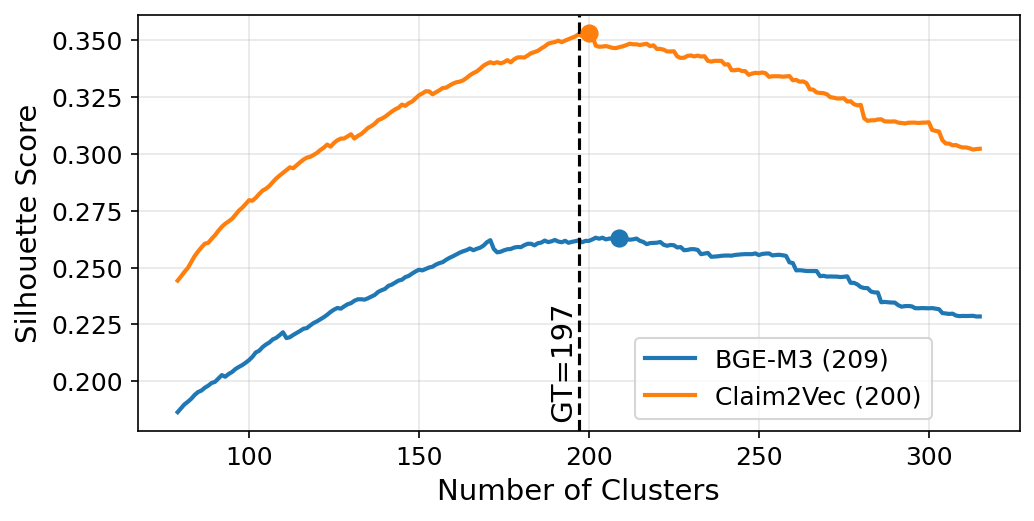}
    \caption{SS - \textit{ClaimCheck}}
    \label{fig:silhouette_comparison_FactCheck}   
    \end{subfigure}  
    \begin{subfigure}[b]{0.3\textwidth}
    \includegraphics[width=\textwidth]{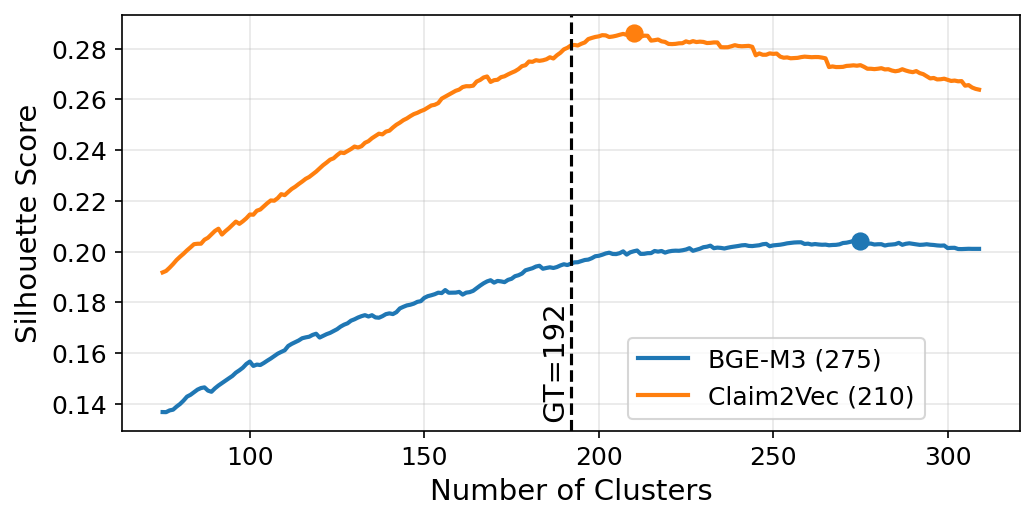}
    \caption{SS - \textit{ClaimMatch}}
    \label{fig:silhouette_comparison_ClaimMatching}   
    \end{subfigure}
    \begin{subfigure}[b]{0.32\textwidth}
    \includegraphics[width=\textwidth]{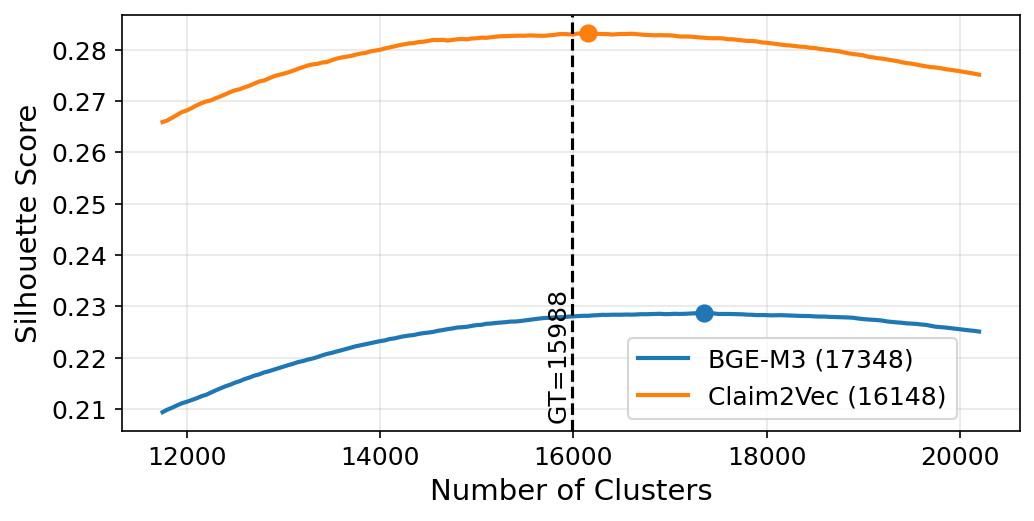}
    \caption{SS - \textit{MultiClaim-Test}}
    \label{fig:silhouette_comparison_MultiClaim_1}   
    \end{subfigure}
    \caption{ARI, AMI, and Silhouette Score (SS) vs Number of Clusters Produced}
    \label{fig:cluster_configuration_analysis}
\end{figure*}

\begin{table}[t]
\centering
\scriptsize
\renewcommand{\arraystretch}{0.9}
\begin{tabular}{lcccc}
\hline
 & \multicolumn{2}{c}{Split} & \multicolumn{2}{c}{Mismerge} \\
\hline
Dataset & BGE-M3 & Claim2Vec & BGE-M3 & Claim2Vec \\
\hline
ClaimCheck      & 57   & 37   & 26   & 18   \\
ClaimMatch      & 146  & 83   & 48   & 37   \\
MultiClaim-Test & 8,474 & 6,383 & 3,495 & 3,258 \\
\hline
\end{tabular}
\caption{Number of split and mismerge errors produced by \textit{BGE-M3} and \textit{Claim2Vec} across datasets.}
\label{tab:split_mismerge}
\end{table}

We further analyze two types of clustering errors: split errors, where instances from the same ground-truth cluster are assigned to different predicted clusters, and mismerge errors, where instances from different ground-truth clusters are grouped into the same predicted cluster. Table \ref{tab:split_mismerge} shows that \textit{Claim2Vec} reduces split errors more substantially than mismerge errors, with improvements growing for more complex datasets. This suggests fine-tuning primarily consolidates semantically similar claims in the embedding space, effectively pulling positive claims closer together while still controlling the rate of incorrect merges. Appendix \ref{app:split_mismerge} provides visualizations of both error types, along with illustrative examples of cluster splits that are corrected by \textit{Claim2Vec} across datasets.

\begin{table}[t]
\centering
\scriptsize
\setlength{\tabcolsep}{4pt}
\begin{tabular}{llccccc}
\hline
\multirow{2}{*}{Dataset} & \multirow{2}{*}{Model} 
& \multicolumn{3}{c}{AUC} 
& \multicolumn{2}{c}{Maximum} \\
\cline{3-7}
& & ARI & AMI & SS 
& ARI & AMI \\
\hline

\multirow{2}{*}{ClaimCheck} 
& BGE-M3 
& 0.776 & 0.891 & 0.242 
& 0.876 & 0.940 \\
& Claim2Vec 
& \textbf{0.803} & \textbf{0.902} & \textbf{0.320} 
& \textbf{0.908} & \textbf{0.954} \\
\hline

\multirow{2}{*}{ClaimMatch} 
& BGE-M3 
& 0.668 & 0.826 & 0.186 
& 0.746 & 0.863\\
& Claim2Vec 
& \textbf{0.693} & \textbf{0.846} & \textbf{0.259} 
& \textbf{0.762} & \textbf{0.885}\\
\hline

\multirow{2}{*}{MultiClaim-Test} 
& BGE-M3 
& 0.590 & 0.718 & 0.224 
& 0.611 & 0.729\\
& Claim2Vec 
& \textbf{0.608} & \textbf{0.747} & \textbf{0.279} 
& \textbf{0.629} & \textbf{0.758}\\
\hline

\end{tabular}
\caption{AUC and Maximum of ARI, AMI and SS for Different Cluster Configurations.}
\label{tab:auc_and_max}
\end{table}

\begin{figure*}[t]
\centering    
\includegraphics[width=0.98\textwidth]{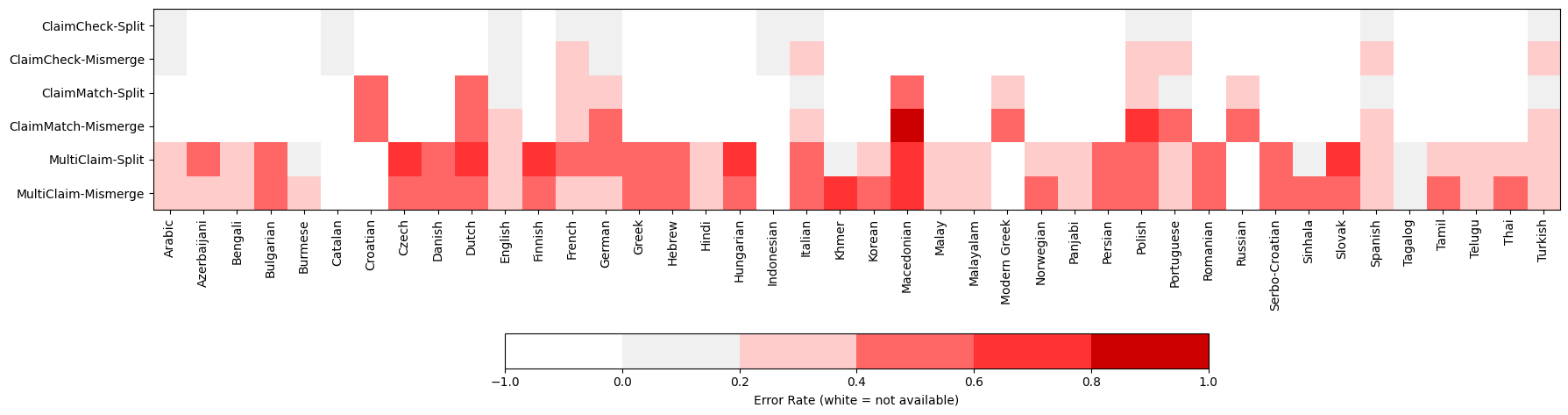}
\caption{Heatmap of Split and Merge Error Rates Across Languages in \textit{Claim2Vec}}
\label{fig:Language-wise-error-analysis}
\end{figure*}

\begin{figure}[t]
    \centering
    \begin{subfigure}[b]{0.23\textwidth}
    \includegraphics[width=\textwidth]{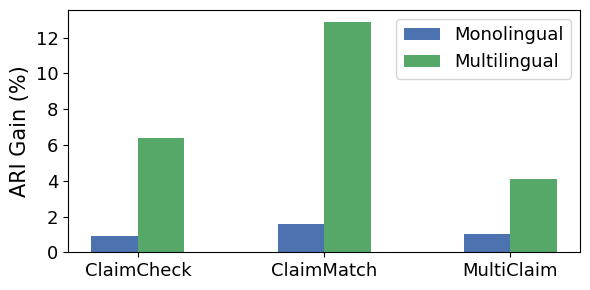}
    \caption{ARI Gain}
    \label{fig:ARI_Gain}   
    \end{subfigure}  
    \begin{subfigure}[b]{0.23\textwidth}
    \includegraphics[width=\textwidth]{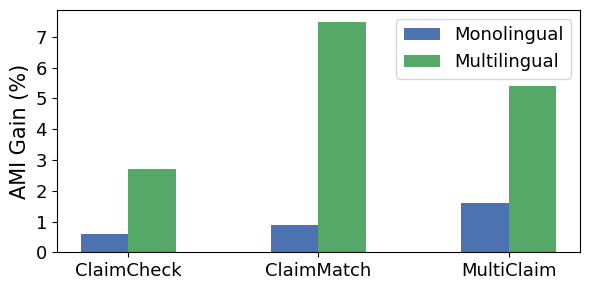}
    \caption{AMI Gain}
    \label{fig:AMI_Gain}   
    \end{subfigure}
    \begin{subfigure}[b]{0.23\textwidth}
    \includegraphics[width=\textwidth]{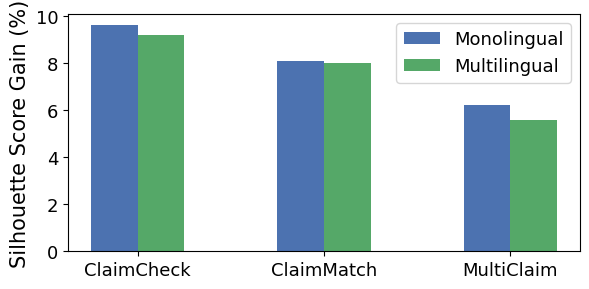}
    \caption{Silhouette Score Gain}
    \label{fig:SS_Gain}   
    \end{subfigure}
    \caption{Multilingual vs Monolingual Gain (\%)}
    \label{fig:gain_analysis}
\end{figure}

\subsubsection{Robust Clustering}
Although we report clustering results using the configuration automatically selected via maximum Silhouette Score, a single clustering state does not fully capture the impact of  embedding models. To assess robustness, we therefore analyze performance across a range of cluster configurations. Specifically, we evaluate configurations within $\pm 10\%$ of the ground-truth number of clusters. Since \textit{MultiClaim-Test} is substantially larger, we vary the number of clusters with a step size of 50 to ensure computational feasibility.

Figure \ref{fig:cluster_configuration_analysis} presents ARI, AMI, and Silhouette Score for varying cluster configurations, with their respective maxima highlighted. Across configurations, \textit{Claim2Vec} consistently yields superior clustering performance, as reflected by the larger area under the curves. The improvement is particularly pronounced for the Silhouette Score, indicating stronger geometric separation in the embedding space. With the exception of ARI and AMI on \textit{ClaimMatch}, the optimal values generally occur near ground-truth cluster counts for \textit{Claim2Vec}. Notably, for \textit{Claim2Vec}, the maximum Silhouette Score closely aligns with the maximum ARI and AMI, suggesting that geometric separation in the embedding space strongly correlates with accurate label assignments. This observation further validates our Silhouette-based strategy for configuration selection. Table \ref{tab:auc_and_max} reports the area under the curve (AUC) for each metric, along with the maximum ARI and AMI.

\subsubsection{Multilingual Performance}
We assess fine-tuning with multilingual data by analyzing clusters that contain claims written in more than one language. Specifically, we compute the performance gain—defined as the difference between \textit{Claim2Vec} and \textit{BGE-M3}—for two cluster types: (i) monolingual clusters, which contain claims in a single language, and (ii) multilingual clusters, which contain claims in multiple languages. Figure \ref{fig:gain_analysis} presents the gain in ARI, AMI, and Silhouette Score across all three datasets. Notably, the improvements in ARI and AMI are substantially higher for multilingual clusters, suggesting effective cross-lingual knowledge transfer during fine-tuning. In contrast, Silhouette Score gains occur for both types, indicating that fine-tuning enhances the global geometric structure of the embedding space irrespective of language composition.  

Although multilingual clusters benefit substantially from the fine-tuning process, we observe that performance varies across languages in all three datasets. Figure \ref{fig:Language-wise-error-analysis} presents the language-wise split and mismerge rates, computed as the fraction of errors relative to the total number of occurrences of each language in the dataset in clusters obtained using \textit{Claim2Vec}. Notably, high- and medium-resource languages such as Arabic, English, French, Italian, Portuguese, Spanish, and Turkish consistently exhibit lower error rates across datasets. In contrast, several languages—including Macedonian, Dutch, Bulgarian, Croatian, Czech, Danish, Finnish, Greek, Hebrew, Persian, Romanian, and Slovak—demonstrate persistently higher error rates, regardless of dataset or error type. This disparity may stem from comparatively weaker representation of these languages in the base encoder’s pretraining data and the multilingual fine-tuning corpus. Interestingly, certain languages appear to be associated predominantly with one type of error. For example, Azerbaijani primarily results in split errors, whereas Korean, Modern Greek, Norwegian, Russian, Sinhala, Tamil, and Thai are mainly associated with mismerge errors. These patterns may reflect language-specific linguistic characteristics—such as morphology, syntax, or semantic variability—that influence how claims are represented and separated in the embedding space.

\section{Conclusion}
This paper introduces \textit{Claim2Vec}, the first multilingual claim embedding model designed to enhance the representation of semantically similar fact-checked claims for clustering. We fine-tune a pretrained multilingual encoder using similar multilingual claim pairs under a contrastive learning framework. The resulting model, \textit{Claim2Vec}, demonstrates consistently strong performance on claim clustering, outperforming 14 multilingual embedding models across three datasets that vary in size and density. In particular, \textit{Claim2Vec} corrects the majority of split errors produced by the base pretrained encoder, \textit{BGE-M3}, while significantly improving alignment with ground-truth cluster assignments and enhancing geometric separation in the embedding space. Our analysis shows that these improvements remain consistent across different cluster configurations, underscoring the robustness of \textit{Claim2Vec} as a claim representation model. A detailed multilingual analysis reveals that clusters containing claims in multiple languages benefit substantially from fine-tuning, indicating effective cross-lingual knowledge transfer. Future work will investigate the impact of \textit{Claim2Vec} on other components of automated fact-checking systems, particularly retrieval of previously fact-checked claims from multilingual claim databases. 

\section*{Limitations}
While advancing research in an underexplored task such as claim clustering to support automated fact-checking, we acknowledge the following limitations in our work:
\begin{compactitem}
    \item To remain consistent with the datasets used in our experiments, we assume that each claim expresses a single factual statement. In practice, however, claims may contain multiple factual assertions. Extending the clustering framework to effectively handle claims covering multiple facts remains an important direction for future work.
    
    \item We evaluate the impact of fine-tuning using only \textit{BGE-M3} as the pretrained encoder. Although this model performs strongly compared to other multilingual embedding models, the generalizability of our approach could be further strengthened by demonstrating fine-tuning with additional multilingual encoders.

    \item Despite the extensive and diverse set of 86 languages included in the MultiClaimNet benchmark dataset that we employ for our research, broader and more comprehensive study of multilingual claim clustering datasets is beyond the scope of our work, primarily due to the lack of more resources and datasets.
\end{compactitem}

\section*{Acknowledgments}
This project is funded by the European Union and UK Research and Innovation under Grant No. 101073351 as part of Marie Skłodowska-Curie Actions (MSCA Hybrid Intelligence to monitor, promote, and analyze transformations in good democracy practices). We acknowledge Queen Mary's Apocrita HPC facility, supported by QMUL Research-IT, for enabling our experiments \cite{king_2017_438045}.

\bibliography{references}

\appendix

\section{\textit{MultiClaim} Train \& Test Partition Topics}
\label{app:multilclaim_topics}

\begin{figure}[h]
\centering    
\includegraphics[width=0.45\textwidth]{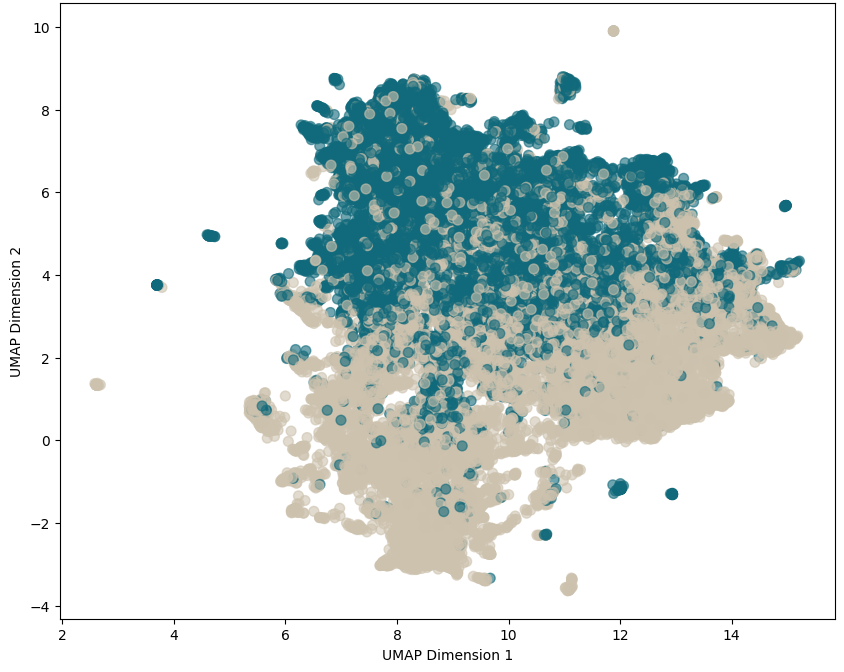}
\caption{2D Projection of Two Topic Groups in \textit{MultiClaim}}
\label{fig:topic_groups_2D}
\end{figure}

\begin{figure*}[t]
    \centering
    \begin{subfigure}[b]{0.46\textwidth}
    \includegraphics[width=\textwidth]{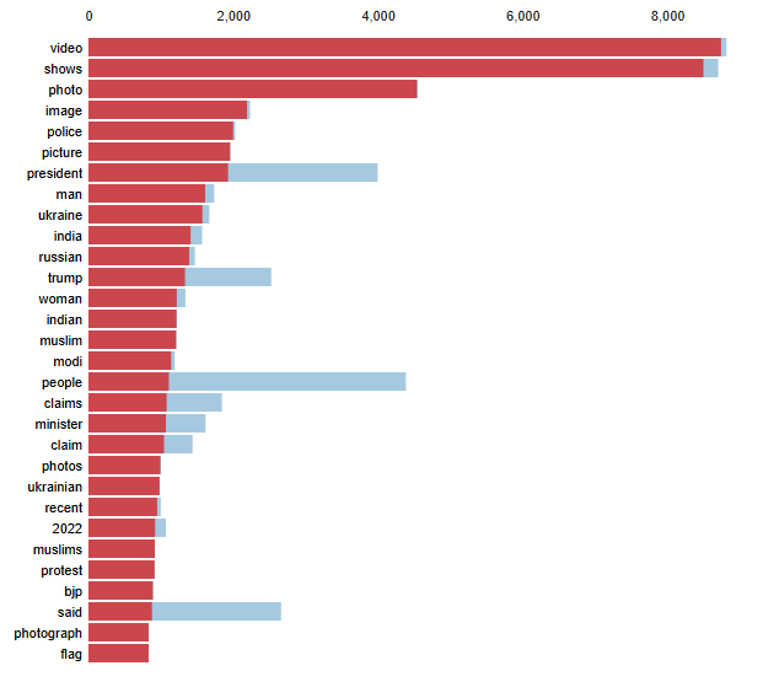}
    \caption{Topic Group 1}
    \label{fig:Topic_2_words}   
    \end{subfigure}  
    \begin{subfigure}[b]{0.47\textwidth}
    \includegraphics[width=\textwidth]{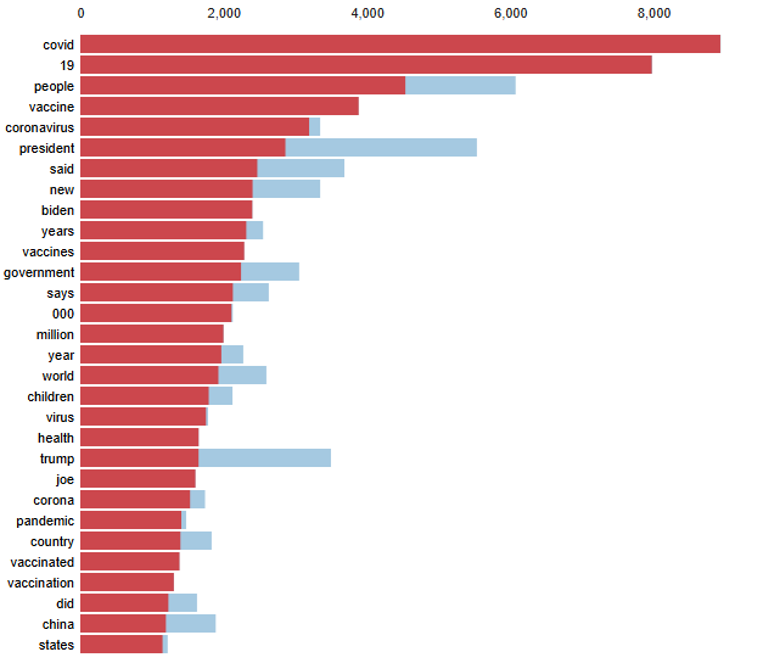}
    \caption{Topic Group 2}
    \label{fig:Topic_1_words}   
    \end{subfigure}
    \begin{subfigure}[]{0.47\textwidth}
    \includegraphics[width=\textwidth]{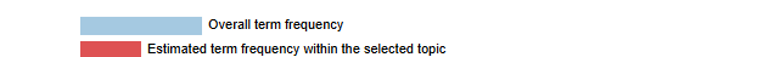}
    \end{subfigure}
    \caption{Top 30 Frequent Words of Two Topic Groups in \textit{MultiClaim}}
    \label{fig:topics_word_distribution}
\end{figure*}

Figure \ref{fig:topic_groups_2D} shows the 2D projection of claims belonging to the two topic groups in \textit{MultiClaim}. Although the groups were originally identified using the topic model Latent Dirichlet Allocation, a clear separation between them is also visible in the embedding space. Figures \ref{fig:Topic_2_words} and \ref{fig:Topic_1_words} present the distributions of the top 30 words associated with each topic group. Topic group 1 primarily discusses recent political issues, whereas topic group 2 mainly focuses on COVID-19–related discussions. This indicates a clear topical separation between the training and test partitions.

\subsection{Hyperparameter Search}\label{app:hyperparamters}

Table \ref{tab:hyperparameters} reports the hyperparameter search results based on the optimal Silhouette Score. As noted earlier, for \textit{MultiClaim-Test} the hyperparameters are determined as the average over five random samples of 10,000 records each due to computational constraints.

\begin{table*}[!t]
\scriptsize
\centering
\begin{tabular}{llllll}
\hline
Approach                       & Hyperparameter            & Search Options                        & ClaimCheck & ClaimMatch & MultiClaim-Test \\ \hline
\multirow{4}{*}{HDBSAN}        & vector dimension         & 2-10    & 4 & 4 & 8      \\ 
                               & cluster selection epsilon & [0-0.1], step size of 0.01                         & 0.1 & 0.1 & 0.055    \\ 
                               & min samples               & Minimum value                         & 1 & 1 & 1       \\ 
                               & min cluster size          & Minimum value                         & 2 & 2 & 2       \\ \hline
\multirow{2}{*}{Agglomerative} & distance threshold         & {[}0.5 - 1.5{]}, step size of 0.05       & 1.38 & 1.258 & 0.95       \\ 
                               & linkage                   & Default & ward & ward & ward    \\ \hline
Birch                          & threshold                 & {[}0.1 - 1{]}, step size 0.1          & 0.5 & 0.5 & 0.44     \\
& branching factor & {[}10 - 100{]}, step size 10          & 60 & 70 & 66     \\\hline
Optics & vector dimension         & 2-10    & 4 & 8 & 8      \\ 
                               & cluster selection epsilon & [0-0.1], step size of 0.01                         & 0.1 & 0.1 & 0.1    \\ 
                               & min samples               & Minimum value                         & 2 & 2 & 2       \\ 
                               & min cluster size          & Minimum value                         & 2 & 2 & 2       \\
                               \hline \\
                               Meanshift                      & bandwidth                 & {[}0.5 - 1.0{]}, step size of 0.05    & 0.7 & 0.7 & -    \\ \hline
Affinity Propagation                      & damping                 & {[}0.5 - 0.95{]}, step size of 0.05    & 0.5 & 0.65 & -    \\ \hline
Leiden Community Detection                      & nearest neighbors                 & [2-5]    & 3 & 3 & 2    \\ 
& resolution                 & [5-20]    & 5 & 6 & 19    \\\hline
\end{tabular}
\caption{Hyperparameters of Clustering Approaches}\label{tab:hyperparameters}
\end{table*}

\section{Split vs Mismerge}\label{app:split_mismerge}

\begin{figure}[h]
    \centering
    \begin{subfigure}[b]{0.23\textwidth}
    \includegraphics[width=\textwidth]{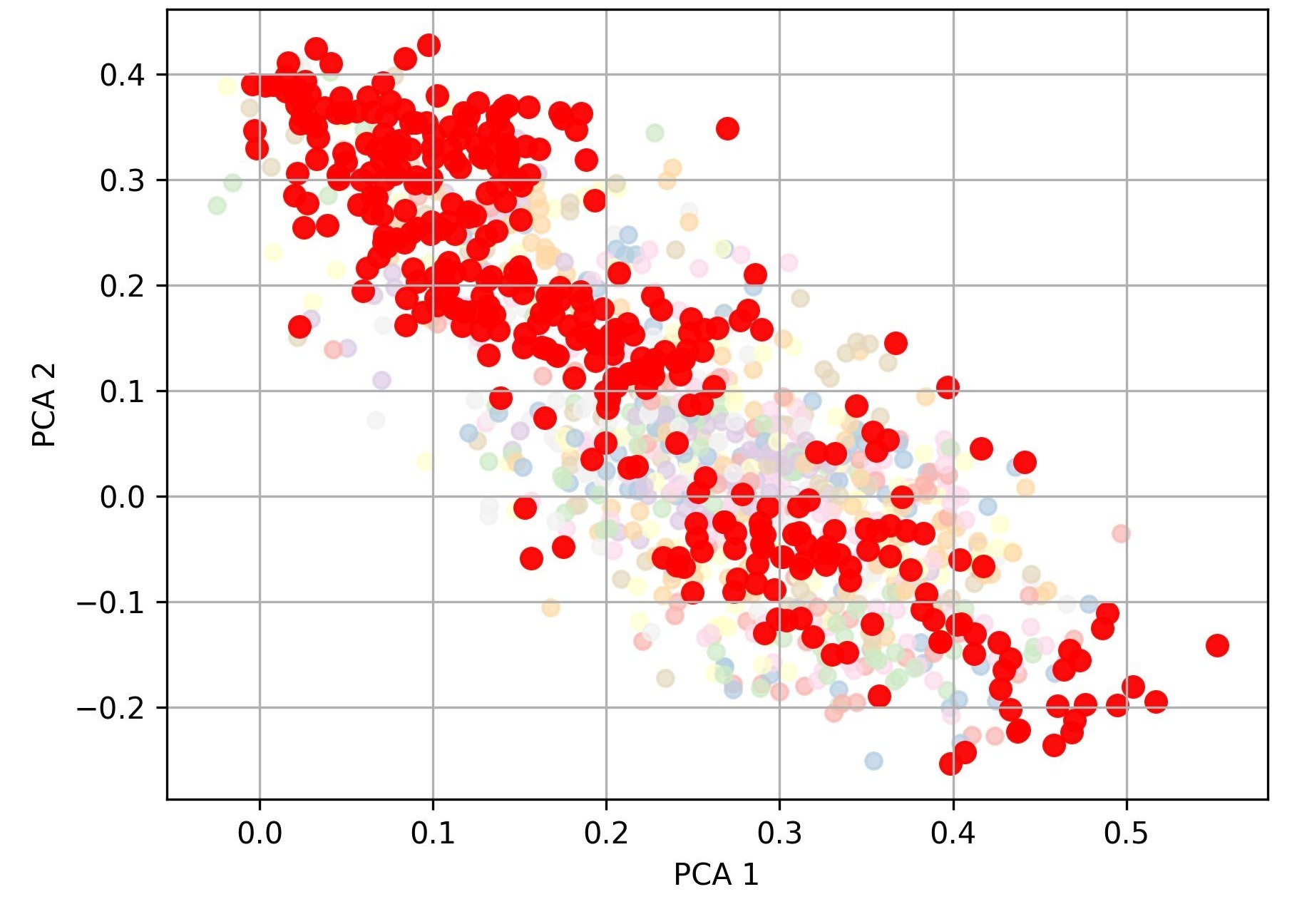}
    \caption{57 Split - \textit{BGE-M3}}
    \label{fig:split_clusters_claimcheck_bgem3}   
    \end{subfigure} 
    \begin{subfigure}[b]{0.23\textwidth}
    \includegraphics[width=\textwidth]{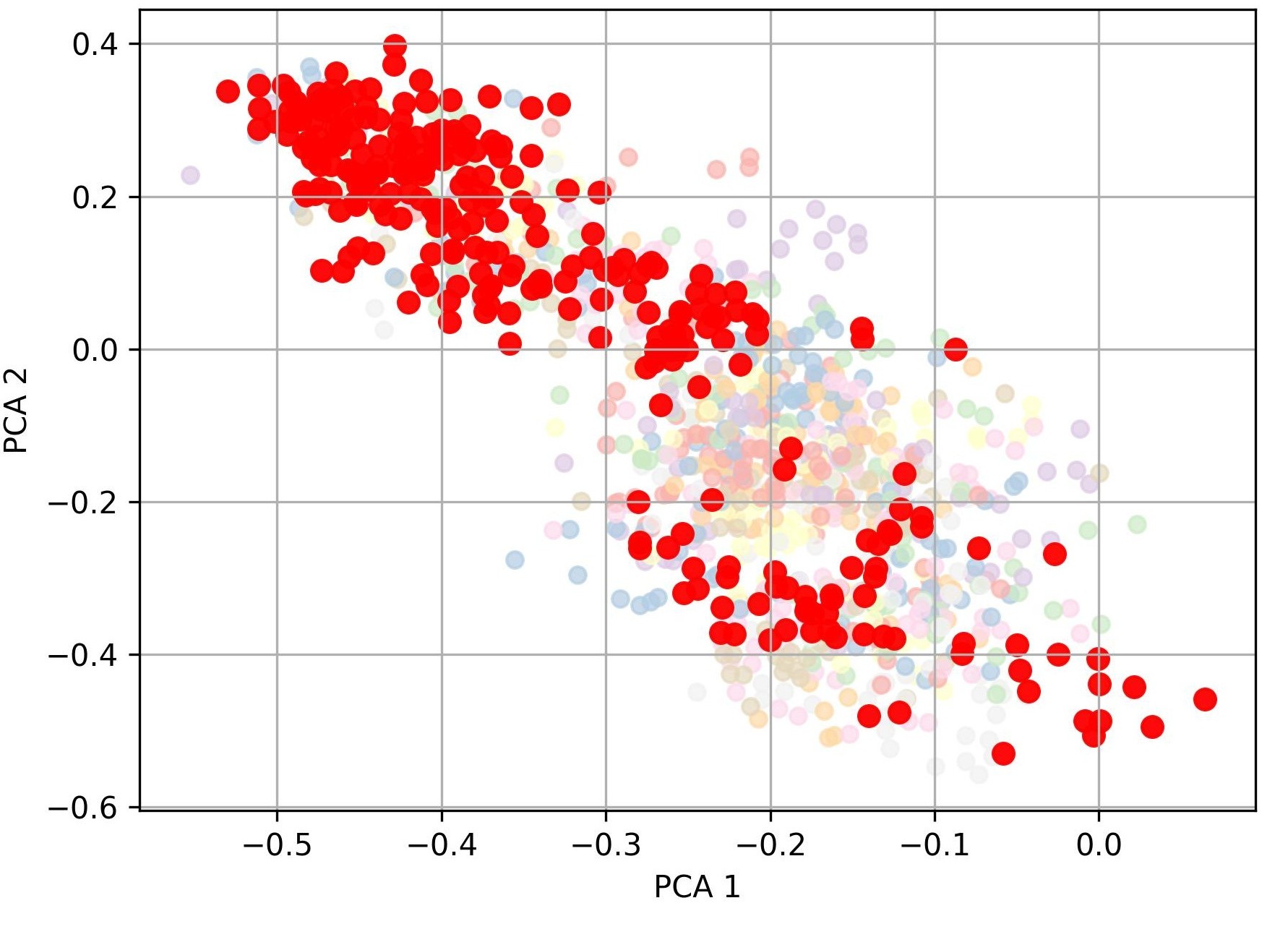}
    \caption{37 Split - \textit{Claim2Vec}}
    \label{fig:split_clusters_claimcheck_claim2vec}   
    \end{subfigure}\\
    \begin{subfigure}[b]{0.23\textwidth}
    \includegraphics[width=\textwidth]{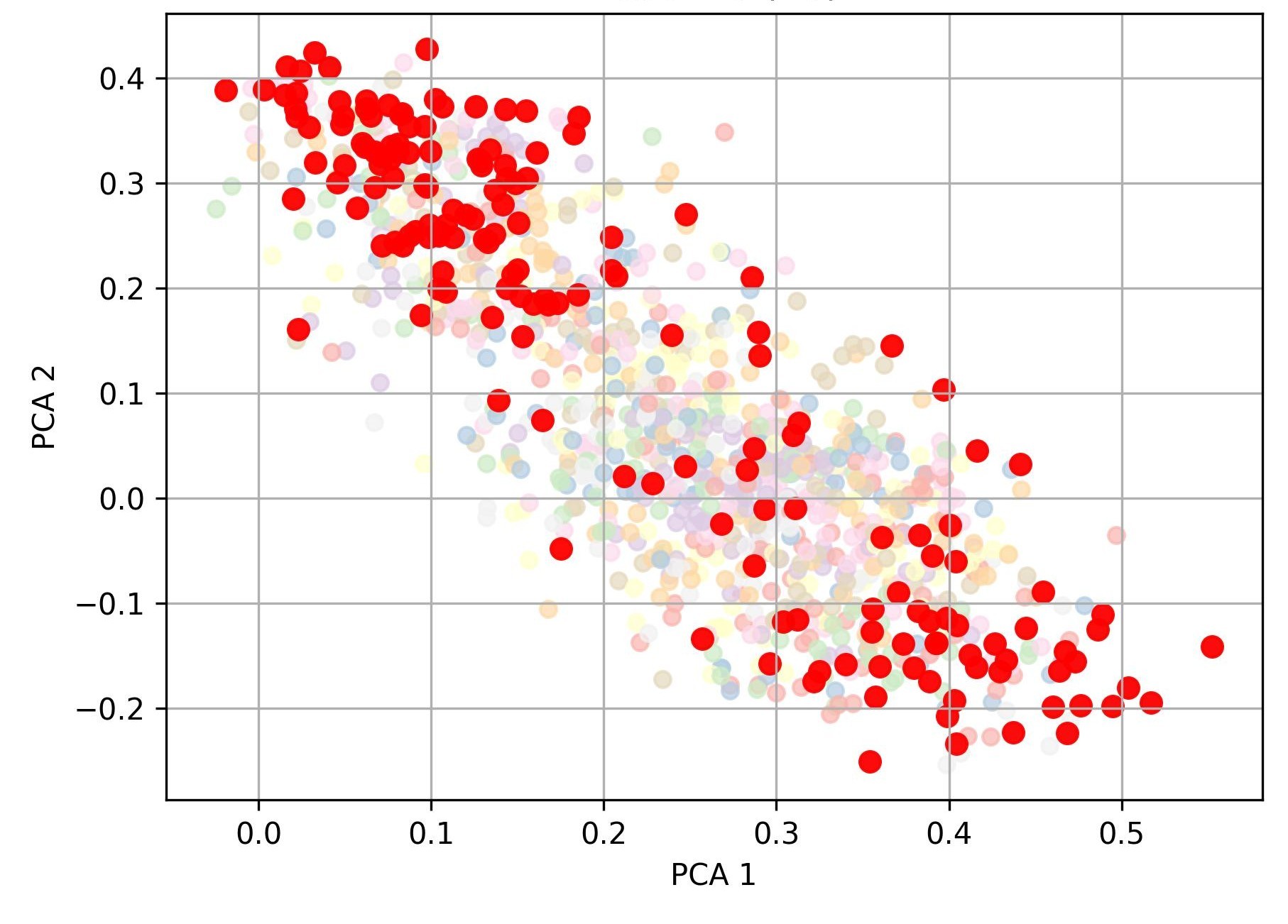}
    \caption{26 Mismerge - \textit{BGE-M3}}
    \label{fig:mismerged_clusters_claimcheck_bgm3}   
    \end{subfigure}
    \begin{subfigure}[b]{0.23\textwidth}
    \includegraphics[width=\textwidth]{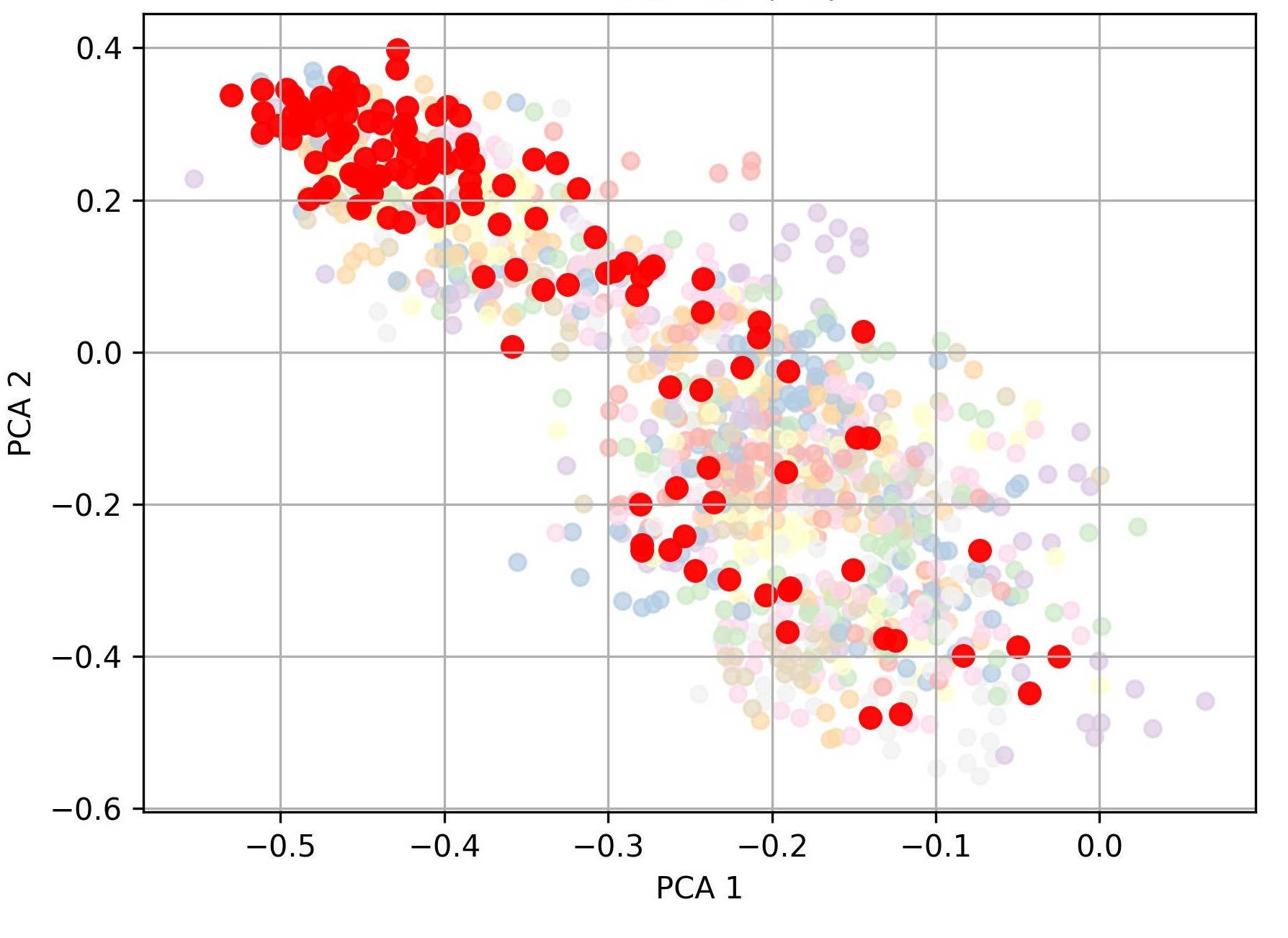}
    \caption{18 Mismerge - \textit{Claim2Vec}}
    \label{fig:mismerged_clusters_claimcheck_claim2vec}   
    \end{subfigure}
    \caption{2D projection of claims in \textit{ClaimCheck} with the red color data points highlighting claims belonging to the split and mismerge error types}
    \label{fig:split_mismerge_claimcheck}
\end{figure}

\begin{figure}[h]
    \centering
    \begin{subfigure}[b]{0.23\textwidth}
    \includegraphics[width=\textwidth]{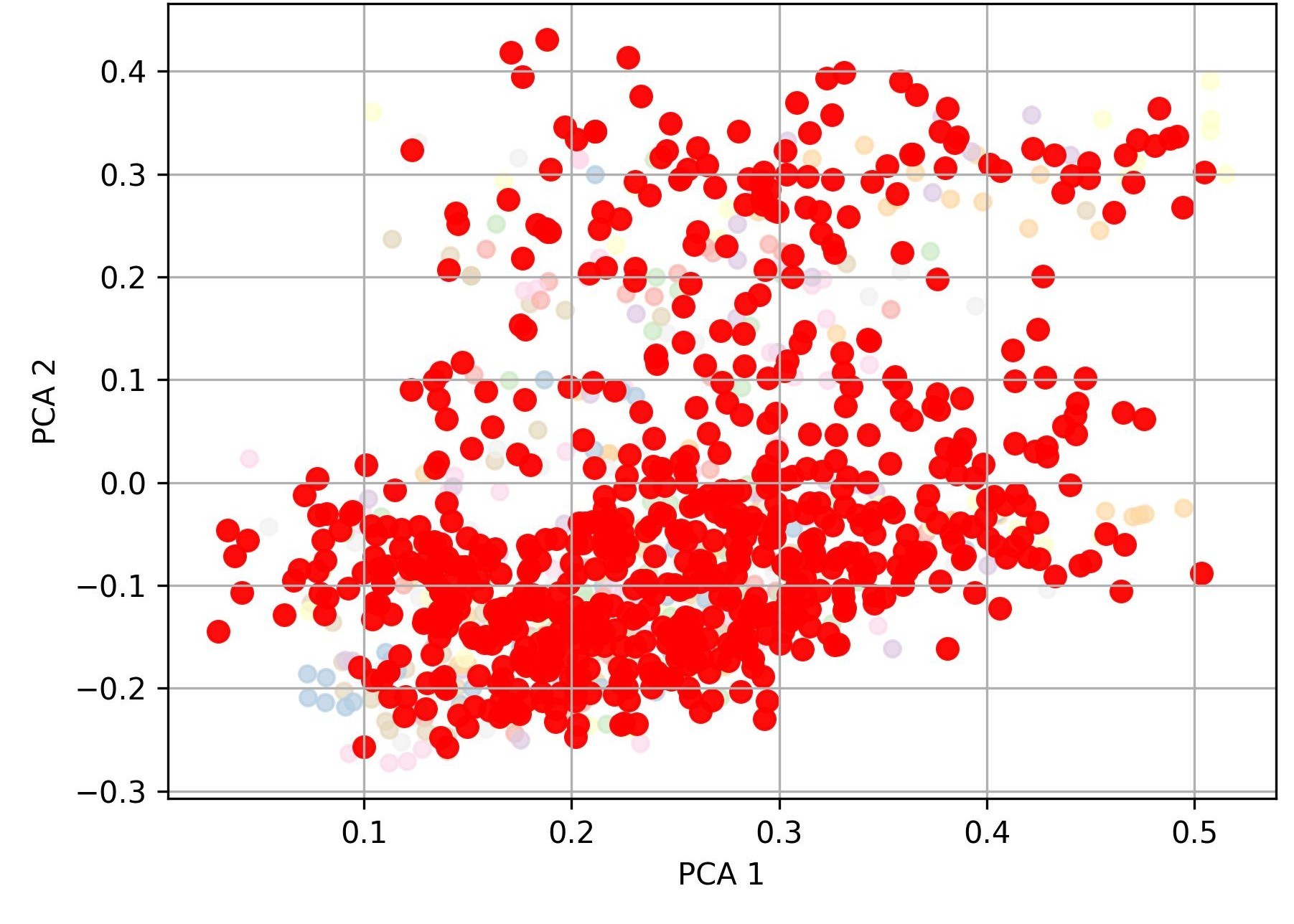}
    \caption{146 Split - \textit{BGE-M3}}
    \label{fig:split_clusters_claimmatch_bgem3}   
    \end{subfigure}  
    \begin{subfigure}[b]{0.23\textwidth}
    \includegraphics[width=\textwidth]{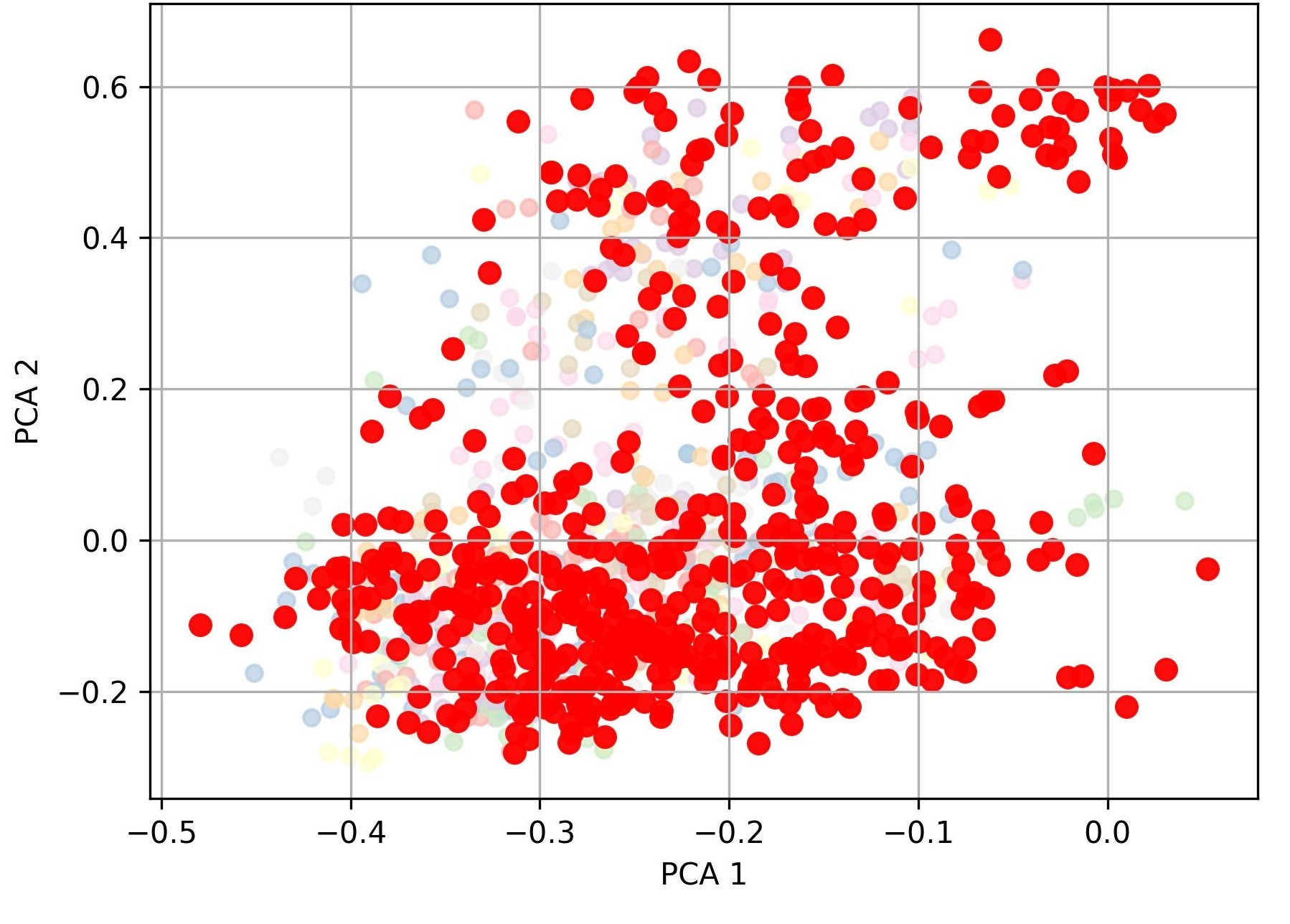}
    \caption{83 Split - \textit{Claim2Vec}}
    \label{fig:split_clusters_claimmatch_claim2vec}   
    \end{subfigure}\\
    \begin{subfigure}[b]{0.23\textwidth}
    \includegraphics[width=\textwidth]{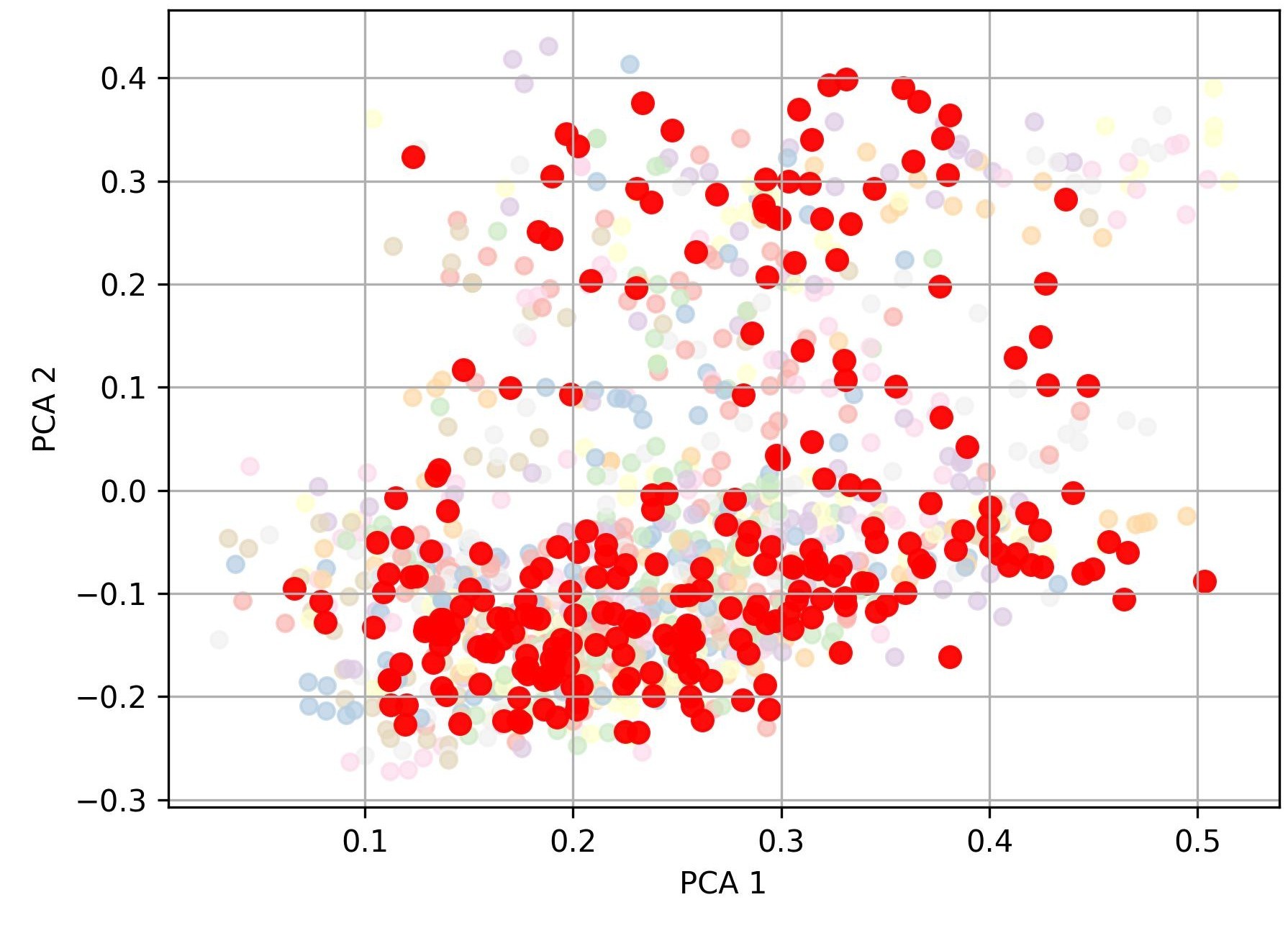}
    \caption{48 Mismerge - \textit{BGE-M3}}
    \label{fig:mismerged_clusters_claimmatch_bgem3}   
    \end{subfigure}
    \begin{subfigure}[b]{0.23\textwidth}
    \includegraphics[width=\textwidth]{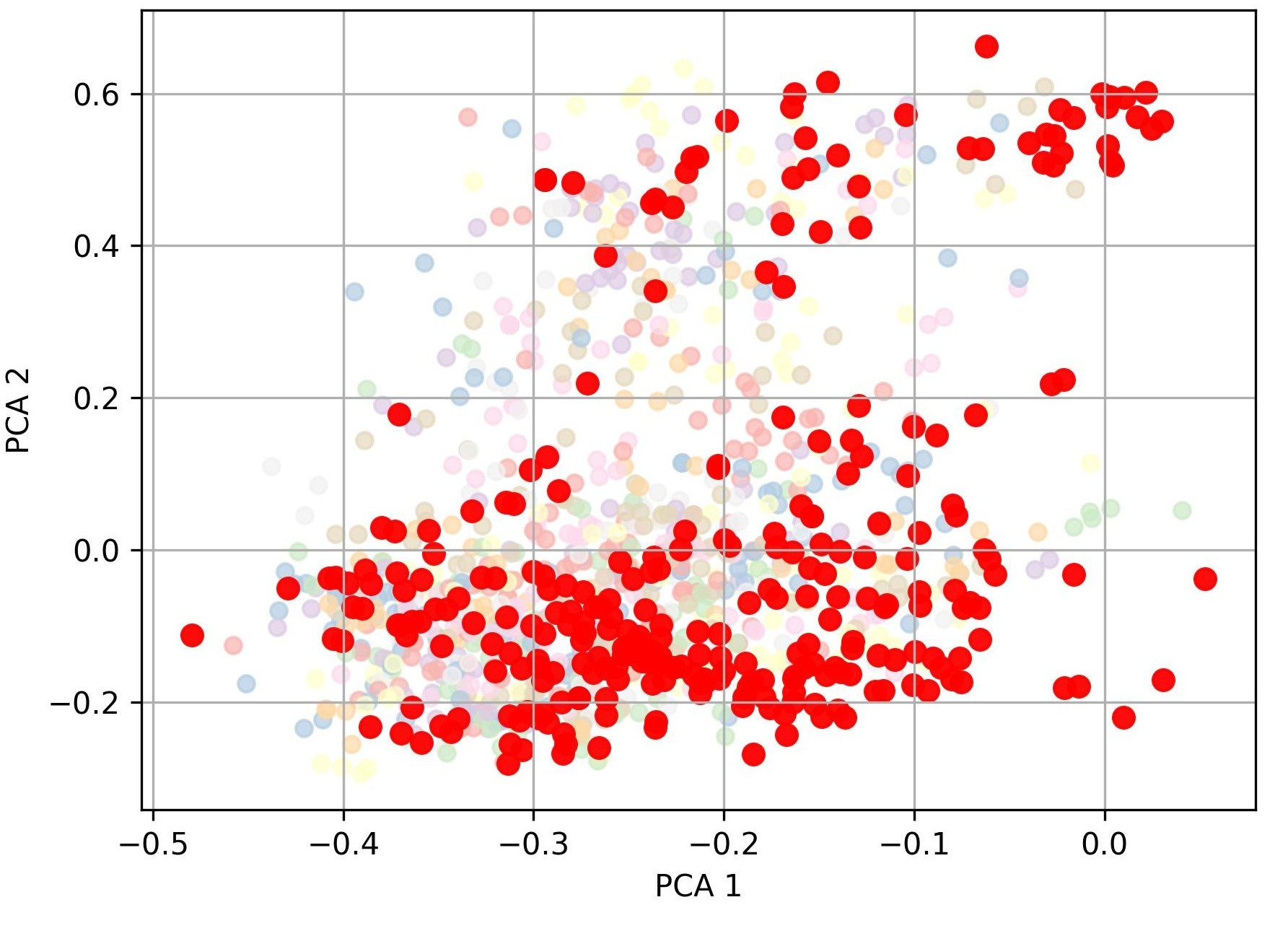}
    \caption{37 Mismerge - \textit{Claim2Vec}}
    \label{fig:mismerged_clusters_claimmatch_claim2vec}   
    \end{subfigure}  
    \caption{2D projection of claims in \textit{ClaimMatch} with the red color data points highlighting claims belonging to the split and mismerge error types}
    \label{fig:split_mismerge_claimmatch}
\end{figure}

Figures \ref{fig:split_mismerge_claimcheck} and \ref{fig:split_mismerge_claimmatch} show the 2D projections of claims in \textit{ClaimCheck} and \textit{ClaimMatch}, respectively, along with the data points contributing to split and mismerge errors produced by \textit{BGE-M3} and \textit{Claim2Vec} during clustering. As discussed earlier, \textit{Claim2Vec} reduces the number of incorrect splits more substantially than mismerge errors.

\begin{table*}[t]
\centering
\small
\begin{tabular}{p{12cm}cc}
\hline
\textbf{Claim (Language)} & \textbf{BGE-M3} & \textbf{Claim2Vec} \\
\hline
COVID-19 vaccines contain graphene oxide (es) & \cellcolor{red!30}53 & \cellcolor{blue!30}2 \\
Covid-19 vaccines contain graphene oxide (es) & \cellcolor{red!30}53 & \cellcolor{blue!30}2 \\
COVID-19 vaccines contain graphene oxide. (fr) & \cellcolor{red!30}53 & \cellcolor{blue!30}2 \\
Pfizer and Moderna COVID-19 vaccines contain toxic graphene oxide (es) & \cellcolor{red!30}53 & \cellcolor{blue!30}2 \\
Covid19 vaccines have graphene oxide (es) & \cellcolor{red!30}53 & \cellcolor{blue!30}2 \\
Covid vaccines contain graphene (es) & \cellcolor{red!30}53 & \cellcolor{blue!30}2 \\
Graphene oxide is part of the COVID-19 vaccine (ru) & \cellcolor{red!30}53 & \cellcolor{blue!30}2 \\
Graphene oxide is found in [covid-19] vaccines, it is irrefutable (es) & \cellcolor{red!30}53 & \cellcolor{blue!30}2 \\
Covid-19 vaccines contain graphene hydroxide (es) & \cellcolor{red!30}53 & \cellcolor{blue!30}2 \\
The content [of the covid-19 vaccines] is mostly graphene oxide (es) & \cellcolor{red!30}53 & \cellcolor{blue!30}2 \\
Covid-19 vaccines contain graphene oxide (es) & \cellcolor{red!30}53 & \cellcolor{blue!30}2 \\
COVID vaccine contains graphene oxide (es) & \cellcolor{red!30}53 & \cellcolor{blue!30}2 \\
Vaccines contain graphene oxide (it) & \cellcolor{green!30}54 & \cellcolor{blue!30}2 \\
Vaccines contain graphene oxide. (fr) & \cellcolor{green!30}54 & \cellcolor{blue!30}2 \\
Vaccines contain graphene hydroxide blades that cut blood vessels (es) & \cellcolor{green!30}54 & \cellcolor{blue!30}2 \\
Vaccines contain graphene oxide. These vaccines weaken the immune system. (es) & \cellcolor{green!30}54 & \cellcolor{blue!30}2 \\
Vaccines on the market contain carcinogenic graphene oxide (en) & \cellcolor{green!30}54 & \cellcolor{blue!30}2 \\
Pfizer's vaccine contains graphene oxide (fr) & \cellcolor{green!30}54 & \cellcolor{blue!30}2 \\
Vaccines contain Graphene Oxide (fr) & \cellcolor{green!30}54 & \cellcolor{blue!30}2 \\
Vaccines contain graphene that causes clots in the brain (es) & \cellcolor{green!30}54 & \cellcolor{blue!30}2 \\
Former Pfizer employee confirms vaccine contains graphene oxide nanoparticles (es) & \cellcolor{yellow!30}55 & \cellcolor{blue!30}2 \\
A Facebook post claimed that Spanish doctors had detected graphene oxide in Covid vaccines. (mk) & \cellcolor{yellow!30}55 & \cellcolor{blue!30}2 \\
Video of an autopsy of the brain of a person who has received the COVID-19 vaccine, which contains graphene (es) & \cellcolor{yellow!30}55 & \cellcolor{blue!30}2 \\
Graphene is being injected into the population. Covid-19 vaccines, PCR swabs contain graphene (es) & \cellcolor{yellow!30}55 & \cellcolor{blue!30}2 \\
A report from the University of Almeria shows that the vaccine against COVID-19 contains graphene oxide (es) & \cellcolor{yellow!30}55 & \cellcolor{blue!30}2 \\
A study by the University of Almeria shows that vaccines contain graphene oxide (es) & \cellcolor{yellow!30}55 & \cellcolor{blue!30}2 \\
ANMAT admitted that vaccines contain graphene (es) & \cellcolor{yellow!30}55 & \cellcolor{blue!30}2 \\
"Youtube censored me! Wrong decision, comrades. And this is IT! We're going to spread it everywhere now! 99\% of the va*cinna*a content is GRAPHENE OXIDE. The video has been shared more than 20,000 times in the Balkans alone! This time we have SCIENTIFIC PROOF, this time the SECRET has been revealed! … Everyone must know about this, even those of you and our friends who have already been 'poked'," reads the post shared with a video taken from StewPeters.tv (mk) & \cellcolor{purple!30}137 & \cellcolor{orange!30}1 \\
\hline
\end{tabular}
\caption{Split error example of claims belonging to the same ground-truth cluster in \textit{ClaimCheck}.}\label{tab:split_example_claimcheck}
\end{table*}
\begin{table*}[t]
\centering
\small
\begin{tabular}{p{12cm}cc}
\hline
\textbf{Claim (Language)} & \textbf{BGE-M3} & \textbf{Claim2Vec} \\
\hline
This video shows a massive full moon appearing in between Russia and Canada in the Arctic (en) & \cellcolor{red!30}215 & \cellcolor{blue!30}73 \\
A video shows a huge Moon appearing in the Arctic Circle (it) & \cellcolor{red!30}215 & \cellcolor{blue!30}73 \\
Video shows eclipse with giant Moon in the Arctic between Russia and Canada! (pt) & \cellcolor{red!30}215 & \cellcolor{blue!30}73 \\
The video shows the moon rising in the Artice circle between Russia and Canada. (en) & \cellcolor{red!30}215 & \cellcolor{blue!30}73 \\
A video shows a gigantic moon moving over the Arctic sky, eclipsing the sun, and disappearing, all in just 30 seconds. (en) & \cellcolor{red!30}215 & \cellcolor{blue!30}73 \\
Video of a giant moon between Russia and Canada (en) & \cellcolor{red!30}215 & \cellcolor{blue!30}73 \\
A video depicts a lunar eclipse in the Arctic region between Russia and Canada. (en) & \cellcolor{red!30}215 & \cellcolor{blue!30}73 \\
Imagine that you are sitting in this place today (between Russia and Canada in the Arctic), when Luna appears in such a large size for 30 seconds, and after blocking the Sun for 5 seconds, it disappears. (ru) & \cellcolor{green!30}216 & \cellcolor{blue!30}73 \\
The Moon can be seen passing "for just 30 seconds" in "all its splendor" between "Canada and Russia" and is "so close that it looks like it's going to collide with Earth". (es) & \cellcolor{green!30}216 & \cellcolor{blue!30}73 \\
In this video we will see the eclipse in the Arctic between Russia and Canada where the Moon made a 30-second path blocking the Sun for 5 seconds (pt) & \cellcolor{green!30}216 & \cellcolor{blue!30}73 \\
"Imagine sitting in this place during the day(in between Russia and Canada in the Arctic) when the moon appears in this big size for 30 seconds and after blocking the Sun for 5 seconds disappears. Glory to God for his creation. \#space \#Moon \#EclipseLunar“ (en) & \cellcolor{green!30}216 & \cellcolor{blue!30}73 \\
This moon video was filmed at the Canada-Alaska-Russia border. This phenomenon occurs only at perigee (it) & \cellcolor{yellow!30}217 & \cellcolor{blue!30}73 \\
"It was filmed within the Arctic Circle, right between the Canada-Alaska-Russia border. It lasts only a few seconds, but the spectacular view is amazing. This phenomenon can only be observed once a year, for 36 seconds. The moon appears in all its glory, then disappears again. It is so close that it seems like it will hit the ground. Immediately after, there is a total solar eclipse lasting 5 seconds during which everything goes black. This phenomenon only occurs at perigee (pt)" & \cellcolor{yellow!30}217 & \cellcolor{blue!30}73 \\
Video filmed in real time. It is the so-called PERIGEE, the phenomenon in which the proximity of the Moon is so evident that we suddenly perceive the great speed at which the Earth moves. (pt) & \cellcolor{yellow!30}217 & \cellcolor{orange!30}74 \\
This video was filmed within the Arctic Circle, precisely between the borders of Canada, Alaska, and Russia. This phenomenon can only be observed once a year, lasting for 36 seconds. The moon appears and disappears. Immediately after that, a total solar eclipse follows, lasting 5 seconds. (en) & \cellcolor{yellow!30}217 & \cellcolor{blue!30}73 \\
It was filmed within the Arctic Circle, right between the Canada-Alaska-Russia border. It lasts only a few seconds, but the spectacular view is remarkable. (pt) & \cellcolor{yellow!30}218 & \cellcolor{blue!30}73 \\
Video of the Moon seen from the Arctic, North Pole, between Russia and Canada, lasts only a few seconds (pt) & \cellcolor{yellow!30}218 & \cellcolor{blue!30}73 \\
This video was filmed within the Arctic Circle, right between the Canada-Alaska-Russia border. It lasts only a few seconds, but the spectacular view is amazing. This phenomenon can only be observed once a year, for 36 seconds. (pt) & \cellcolor{yellow!30}218 & \cellcolor{blue!30}73 \\
VIDEO: The Moon seen from the Arctic, North Pole, between Russia and Canada, only lasts a few seconds, but it's worth it, it's a beautiful sight, it seems to collide with the Earth, but it doesn’t (es) & \cellcolor{yellow!30}218 & \cellcolor{blue!30}73 \\
The claim that the video reflects the movements of the Moon in the Arctic region around Siberia between Russia and Canada (tr) & \cellcolor{purple!30}223 & \cellcolor{blue!30}73 \\
\hline
\end{tabular}
\caption{Split error example of claims belonging to the same ground-truth cluster in \textit{ClaimMatch}.}\label{tab:split_example_claimmatch}
\end{table*}
\begin{table*}[t]
\centering
\small
\begin{tabular}{p{12cm}cc}
\hline
\textbf{Claim (Language)} & \textbf{BGE-M3} & \textbf{Claim2Vec} \\
\hline
Dr. Tasuku Honjo, Professor of Medicine, Nobel Laureate of Japan, said that the Corona virus did not come naturally, but was created by China and released to the world (tel) & \cellcolor{red!30}16039 & \cellcolor{blue!30}14297 \\
Japanese Nobel laureate Dr. Tasuku Honjo said the new coronavirus was engineered in a Chinese laboratory (eng) & \cellcolor{green!30}16042 & \cellcolor{blue!30}14297 \\
Japanese professor, Nobel laureate Tasuku Honjo said that the coronavirus was created artificially in China (rus) & \cellcolor{yellow!30}16043 & \cellcolor{blue!30}14297 \\
Nobel laureate Tasuku Honjo said that the new corona virus is not natural and that it was created in a Chinese laboratory (hbs) & \cellcolor{purple!30}16048 & \cellcolor{blue!30}14297 \\
\hline
\end{tabular}
\caption{Split error example of claims belonging to the same ground-truth cluster in \textit{MultiClaim-test}.}\label{tab:split_example_multiclaim}
\end{table*}

Table \ref{tab:split_example_claimcheck} shows an example of cluster assignments produced by \textit{BGE-M3} and \textit{Claim2Vec} for claims belonging to the same ground-truth cluster in the \textit{ClaimCheck} dataset. \textit{BGE-M3} assigns these claims to four different clusters, whereas \textit{Claim2Vec} misplaces only one claim. This claim is highly noisy, yet \textit{Claim2Vec} still places it in the nearest cluster in the embedding space (cluster IDs 1 and 2). Table \ref{fig:split_clusters_claimmatch_claim2vec} presents a more challenging example from the \textit{ClaimMatch} dataset, which contains relatively longer claims. Again, \textit{BGE-M3} assigns the claims to four different clusters, while \textit{Claim2Vec} misplaces only one claim. This error likely occurs because several key terms (e.g., \textit{Arctic Circle, Canada, Alaska, Russia}) are missing from that claim. Nevertheless, the claim is still placed in a nearby cluster, as reflected by cluster IDs 73 and 74. Table \ref{tab:split_example_multiclaim} shows an example from \textit{MultiClaim-Test}, where \textit{BGE-M3} fails to correctly group claims written in four different languages, resulting in distinct clusters. In contrast, \textit{Claim2Vec} successfully corrects this split, likely due to the cross-lingual knowledge transfer learned during fine-tuning.

\end{document}